%% file: main.tex
\documentclass{article}

\PassOptionsToPackage{numbers, sort&compress}{natbib}
    \usepackage[preprint]{neurips_2025}

\usepackage[utf8]{inputenc} % allow utf-8 input
\usepackage[T1]{fontenc}    % use 8-bit T1 fonts
\usepackage{hyperref}       % hyperlinks
\usepackage{url}            % simple URL typesetting
\usepackage{booktabs}       % professional-quality tables
\usepackage{longtable}
\usepackage{graphicx}
\usepackage{amsfonts}       % blackboard math symbols
\usepackage{nicefrac}       % compact symbols for 1/2, etc.
\usepackage{microtype}      % microtypography
\usepackage{xcolor}         % colors

\usepackage{amsmath}
\usepackage{my_style}
\usepackage{my_commands}
\usepackage{float}
\usepackage{multirow}
\usepackage{array}
\usepackage{subcaption}
\usepackage{svg}

\title{Hard Samples, Bad Labels: Robust Loss Functions That Know When to Back Off}
\author{%
  Nicholas Pellegrino$^{1}$\thanks{Indicates equal contribution, joint first-authorship.}, David Szczecina$^{1,2}$\footnotemark[1], \& Paul Fieguth$^{1}$\\
$^1$Vision and Image Processing Group, Systems Design Engineering, University of Waterloo\\
$^2$Mechanical \& Mechatronics Engineering, University of Waterloo\\
  \texttt{\{npellegr,dszczeci,pfieguth\}@uwaterloo.ca} \\
}

\begin{document}

\maketitle

\begin{abstract} % Limited to one paragraph, 5000 characters in OpenReview

Incorrectly labelled training data are frustratingly ubiquitous in both benchmark and specially curated datasets.
Such mislabelling clearly adversely affects the performance and generalizability of models trained through supervised learning on the associated datasets.
Frameworks for detecting label errors typically require well-trained / well-generalized models; however, at the same time most frameworks rely on training these models on corrupt data, which clearly has the effect of reducing model generalizability and subsequent effectiveness in error detection --- unless a training scheme robust to label errors is employed.
We evaluate two novel loss functions, \textit{Blurry Loss} and \textit{Piecewise-zero Loss}, that enhance robustness to label errors by de-weighting or disregarding difficult-to-classify samples, which are likely to be erroneous.
These loss functions leverage the idea that mislabelled examples are typically more difficult to classify and should contribute less to the learning signal.
Comprehensive experiments on a variety of artificially corrupted datasets demonstrate that the proposed loss functions outperform state-of-the-art robust loss functions in nearly all cases, achieving superior F1 scores for error detection.
Further analyses through ablation studies offer insights to confirm these loss functions' broad applicability to cases of both uniform and non-uniform corruption, and with different label error detection frameworks. 
By using these robust loss functions, machine learning practitioners can more effectively identify, prune, or correct errors in their training data.
Code, including a working demonstration Jupyter Notebook, is available at \url{https://anonymous.4open.science/r/Robust\_Loss-6BAD/}.

% anonymous.4open.science/r/Robust_Loss-6BAD

% We cannot self-plagarize. Below is the abstract from our CVIS paper, so I have written a new abstract above, for NeurIPS
% Methods for detecting label errors in training data require models that are robust to label errors (\ie not fit to erroneously labelled data points). 
% However, acquiring such models often involves training on corrupted data, which presents a challenge.
% Adjustments to the loss function present an opportunity for improvement. 
% Motivated by Focal Loss (which emphasizes difficult-to-classify samples), two novel, yet simple, loss functions are proposed that de-weight or ignore these difficult samples (\ie those likely to have label errors).
% Results on artificially corrupted data show promise, such that F1 scores for detecting errors are improved from the baselines of conventional categorical Cross Entropy and Focal Loss.
\end{abstract}

\section{Introduction}
\label{sec:intro}

Supervised learning relies heavily on the assumption that training labels are accurate; however, label errors are pervasive even in well-established benchmark datasets, typically at rates of at least 5\%~\cite{northcutt2021pervasive}, including in classic datasets such as MNIST~\cite{deng2012mnist,northcutt2021pervasive}.
Label errors (also referred to as label \textit{noise} in the literature) degrade model performance, limit generalizability~\cite{song2022learning,zhang2016understanding,pleiss2020identifying}, and mislead validation metrics.
The prevalence of label errors is especially problematic in domains with hierarchical and fine-grained classes, such as biological datasets~\cite{wu2019ip102,garcin2021pl,van2021benchmarking,he2024species196,nguyen2023insect,gharaee2024bioscan1m,gharaee2024bioscan5m}, where mislabelling often occurs among highly similar categories.
% Further motivation can be found in biological datasets~\cite{wu2019ip102,garcin2021pl,van2021benchmarking,he2024species196,nguyen2023insect,gharaee2024bioscan1m,gharaee2024bioscan5m} where classes are often hierarchical and fine-grained, and distributions are heavy-tailed. 
Efficient automatic error detection could significantly streamline curation and the flagging of samples for expert review, reducing costs and improving dataset quality. 

% Detecting and either pruning (removing) or, ideally, correcting label errors is important because the presence of label errors leads to a reduced ability to train models, meaning accuracy and other key performance metrics are reduced~\cite{song2022learning,zhang2016understanding, ye2023active, pleiss2020identifying}.
% Further, label errors in validation sets mislead practitioners, since the measured validation performance of models does not match their true performance (even ignoring the usual sample biases associated with finite, non-comprehensive sets of validation data)~\cite{northcutt2021pervasive}. 

Existing work on detecting label errors~\cite{jiang2018mentornet,song2019selfie,kim2019nlnl,pleiss2020identifying,northcutt2021confident} such as Confident Learning (CL)~\cite{northcutt2021confident} and Area Under Margin (AUM)~\cite{pleiss2020identifying} typically rely on training surrogate models, 
% (\ie not the model intended for the application utilizing the dataset of question, but rather a separate model, ideally more robust / less prone to overfitting). 
intended to be robust or resistant to overfitting, such that they are well-generalized and not fit to erroneous labels in the training data.
Crucially, the effectiveness of these methods depends on the models producing statistically distinguishable predicted probabilities, $p(k|x)$, for erroneously \vs correctly labelled samples.
However, when trained with standard loss functions such as Cross Entropy~\cite{good1952rational} or Focal Loss~\cite{lin2017focal}, models tend to inadvertently fit to erroneous labels, reducing detection effectiveness. 
% What is desired is a trained model that produces predicted probabilities, $p(k|x)$, for erroneous data that are statistically different than those of correctly labelled samples, such that existing label error detection methods such as CL and AUM can detect this difference and identify erroneous data as such. 
Modification of the loss function used when training such models is one avenue for improving their robustness and downstream utility for detecting label errors.

In this paper, we rigorously investigate two recently introduced robust loss functions ---
 \textit{Blurry Loss} and \textit{Piecewise-zero Loss}~\cite{pellegrino2024loss} --- originally presented with only very few and preliminary studies performed, and specifically designed to improve model robustness by explicitly de-weighting samples likely to be mislabelled.
In datasets with label errors, erroneous samples are likely to be difficult-to-classify.
Inspired by Focal loss~\cite{lin2017focal}, which places emphasis on difficult-to-classify samples, these novel loss functions reverse this idea, assigning less weight to samples with low predicted probability in their as-labelled class (likely to be mislabelled), thereby avoiding detrimental overfitting to erroneous data.
These loss functions result in a more robust training scheme and models that are better suited to detecting label errors than those trained with existing standard or robust loss functions.  

% The concept for the novel loss functions~\cite{pellegrino2024loss} is motivated by Focal loss~\cite{lin2017focal} (FL), which emphasizes difficult-to-classify samples by way of weighing as a function of the predicted probability of the as-labelled class. 
% In datasets with label errors, erroneous samples are likely to be difficult-to-classify. Using conventional Cross Entropy (CE)~\cite{good1952rational} or FL would tend to cause the model to fit to the erroneous data. 
% Rather than emphasize the difficult-to-classify samples, the proposed loss functions de-weight these samples, again, as a function of the predicted probability of the as-labelled class.

Comprehensive experiments are performed to validate these loss functions' effectiveness in improving label error detection, 
% To demonstrate the efficacy of the proposed loss functions, several experiments are performed and briefly outlined here, with further details given in \Cref{sec:experiments}. 
% The overarching experimental framework involves artificially corrupting known error-free (or low error rate) datasets at varying corruption rates, then employing a framework for label detection in which models are trained using the proposed loss functions. 
exploring several artificially corrupted datasets, a range of artificial corruption rates, uniform and non-uniform error statistics, multiple frameworks for label error detection (CL and AUM),
and gradients of corrupted and clean training samples.
% Detections of label errors are measured and compared between models trained using baseline (CE and FL) and existing robust loss functions,  Generalized Cross Entropy~\cite{zhang2018generalized} (GCE) and Active-Negative Loss~\cite{ye2023active} (ANL).
The proposed Blurry and Piecewise-zero loss functions outperform existing loss functions, Generalized Cross Entropy~\cite{zhang2018generalized} (GCE) and Active-Negative Loss~\cite{ye2023active} (ANL), across nearly all experiments, highlighting their utility for practical error detection and correction.

% Lastly, insights on the impact of the proposed loss functions on training dynamics are explored via examination of distributions of the gradients of corrupted and clean training samples. ???]

\section{Preliminaries}
\label{sec:prelim}
Building upon~\cite{zhang2018generalized,ye2023active},
this work is set in the context of a $K$-class classification problem based with $\mathcal{X}\subset\mathbb{R}^r$ as the $r$-dimensional feature space (typically images) and $\mathcal{Y}=\{1,\ldots,K\}$ as the label space (categorical).
A clean (ideal, error-free) dataset is denoted by $D=\{(x_i,y_i)\}_{i=1}^N$ for sets containing $N$ examples and where $(x_i,y_i)\in(\mathcal{X},\mathcal{Y})$.
A classifier is a function, $f:\mathcal{X}\to\mathcal{Y}$, that maps input features to predicted classes by selecting the class, $k$, having the maximal predicted probability, $f=\arg_{k}\max{p(k|x)}$. 
The predicted probabilities, $p(k|x)$, for each class $k\in\mathcal{Y}$, are inferred through a trainable deep neural network (DNN) with a softmax output layer, such that $\sum_{k=1}^K{p(k|x)}=1$. 
In practice, classifiers are trained through the strategy of attempting to minimize the empirical risk, $R_{\mathcal{L}}=\frac{1}{N}\sum_{i=1}^N{\mathcal{L}(p(k|x_i),y_i)}$
% $R_{\mathcal{L}}=\mathbb{E}_D[\mathcal{L}(p(k|x_i),y_i)]=\frac{1}{N}\sum_{i=1}^N{\mathcal{L}(p(k|x_i),y_i)}$
for a given loss function $\mathcal{L}:[0,1]^K\times\mathcal{Y}\to\mathbb{R}^{+}$.

% In the practice of using \textit{active} loss functions~\cite{ma2020normalized}, only the predicted probability, $p(y|x)$, associated with the as-labelled class, $y$, is considered. 
% For example, with Cross Entropy loss, $\text{CE}(p(k|x),y)=-\log(p(y|x))$, the empirical risk to be minimized becomes $R_{\text{CE}}=-\frac{1}{N}\sum_{i=1}^N{\log(p(y_i|x_i))}$.

Label errors are often present within training data, as discussed in \Cref{sec:intro}.
The proportion of data that are mislabelled is termed the corruption rate, $\eta$, and the corrupted dataset is denoted $D_\eta=\{(x_i,\tilde{y_i})\}_{i=1}^N$ for possibly erroneous labels $\tilde{y}_i$.
The probability that a correct label $y$ is mislabelled (corrupted) as label $\hat{y}$ is $\eta_{y\hat{y}}$, such that the corruption rate of specific label $y$ is given by $\eta_{y}=\sum_{\hat{y}\ne{y}}{\eta_{y\hat{y}}}$.
Taking the common assumption that the label corruption (\ie mislabelling) process is conditionally independent of input features~\cite{natarajan2013learning}, $x$, labels are given by
\begin{equation}
    \tilde{y} = 
    \begin{cases}
        y & \text{with probability } (1 - \eta_{y}) \\
        \hat{y},\, \hat{y} \in \mathcal{Y}\setminus\{y\}, & \text{with probability } \eta_{y\hat{y}}
    \end{cases}.
\end{equation}
% The class-dependent corruption rate, $\eta_{y\hat{y}}$, defines a label transition matrix, $T$, such that each entry $T_{y,\hat{y}}$ specifies the probability of a label transition from correct class $y$ to $\hat{y}$, and so $T_{y,\hat{y}} = \eta_{y\hat{y}}$.
% Note that rows of $T$ sum to $1$.
Corruption processes may be categorized based on their conditioning. 
In the most general case of \textit{asymmetric} corruption, the corruption process is class-dependent and conditioned on both correct labels, $y$, and corrupted labels, $\hat{y}$.
% , in which case $T$ is asymmetric. 
Corruption is \textit{symmetric} if the process is \textit{identically} conditioned on both correct and incorrect labels.
% (symmetric corruption), then $T$ is symmetric and both rows and columns of $T$ sum to 1.
Lastly, the corruption process is \textit{uniform} if it is not conditioned on either correct or incorrect labels.
% , then $T$ has all diagonal entries equal, is symmetric, and has all off-diagonal entries also equal.

For the sake of notational simplicity throughout the remainder of this article, predicted probabilities, $p(k|x)$, are stated in terms of an \emph{implied} input $x$, such that the notational simplification is made, from $p(k|x)$ to $p_k$ (indicating predicted probability for class $k$).

\section{Background}
\label{sec:background}
The background information is divided into two main categories: Label Error Detection, and Robust Loss Functions. Because the label error detection frameworks of CL~\cite{northcutt2021confident} and AUM~\cite{pleiss2020identifying} are used in this paper to evaluate the proposed loss functions, more detail is given to them than to other frameworks. 
Similarly, the robust loss functions Generalized Cross Entropy (GCE)~\cite{zhang2018generalized} and Active Negative Loss (ANL)~\cite{ye2023active} are used in this paper and presented in more detail.

\subsection{Label Error Detection}
\label{subsec:label_error_detection}
The prevalence and negative impacts on training models caused by errors in training data have motivated the development of label error detection frameworks. 
Two prominent approaches in this space are Confident Learning (CL)~\cite{northcutt2021confident} and Area Under the Margin (AUM)~\cite{pleiss2020identifying}, both of which rely on training surrogate models on corrupt datasets in order to detect samples likely to have label errors.
Other recent frameworks include MentorNet~\cite{jiang2018mentornet}, SELFIE~\cite{song2019selfie}, and SelNLPL (Selective Negative Learning and Positive Learning)~\cite{kim2019nlnl}.

Confident Learning~\cite{northcutt2021confident} 
% operates under the assumption of an underlying class-conditional noise process, where the probability mislabelling depends only on the latent true class, $y$.
uses k-fold cross-validation to train models on all folds of the mislabelled dataset, wherein, for each fold, the joint distribution of observed, $\tilde{y}$, and underlying true labels, $y$, termed the Confident joint, $Q_{\tilde{y},y}$, is estimated based on the model's predicted probabilities.
 % and counting the number of samples exceeding the expected (average) self-confidence for each class
% By computing the out-of-sample (\ie held-out portion of the dataset for a given fold) predicted probabilities, the framework then identifies examples likely to be mislabelled. 
% Normalized margins are used as a threshold for which likely-to-be-mislabelled examples are detected.
Proposal methods for sample pruning available in the framework include Prune by Class (PBC), Prune by Noise Rate (PBNR), and ``both''.
% (in which the intersection of propositions of PBC and PBNR is used).
PBC detects examples from each class that the model is least confident about, whereas PBNR detects examples that are most likely to be mislabelled as a \textit{different} class.
With PBC, for each class $k$, 
% the number of label errors is estimated by summing the off-diagonal entries of the joint distribution $Q_{\tilde{y}=k,y=j}$ for $k\ne j$. 
% Following this,
examples with the lowest self-confidence (\ie $p(k|x)$, for inputs $x$ labelled class $k$) are proposed for pruning.
With PBNR, 
% for each off-diagonal entry in the estimated joint distribution $Q_{\tilde{y}=k,y=j}$, $k\ne j$, the number of examples that are mislabelled from class $k$ to $j$ are estimated. 
% Following this, 
examples from each class $k$ having the largest margin,
% calculated as the difference between the predicted probability of the example belonging to class $j$ and the predicted probability of the example belonging to its given label $k$ 
(\ie $p(j|x) - p(k|x)$, $j\ne k$), are proposed for pruning.
If the ``both'' method is selected, examples must both be proposed by PBC and PBNR to be detected.
% The framework is model-agnostic; however, accurate probabilistic calibration improves performance.

The Area Under Margin~\cite{pleiss2020identifying} method identifies mislabelled samples based on their training dynamics. 
AUM measures the difference between the logit values for a training example's as-labelled class, $\tilde{y}$, and its highest other class, and averages over the course of training.
Correctly labelled samples tend to exhibit positive AUM values, while mislabelled samples tend to have significantly lower AUM values due to conflicting gradient updates from other samples of the same class.
To determine at which threshold of the AUM statistic samples should be detected as being likely mislabelled, a small portion of the data are re-labelling to a new \textit{fake} class, numbered $K+1$ for a $K$-class problem, and correspondingly one additional output neuron is added to the DNN. The AUM for this set of `threshold samples' is monitored, and the 99th percentile AUM from this set is used as the threshold at which correctly- and mislabelled data are separated.

\subsection{Robust Loss Functions}
The choice of loss function largely determines the training dynamics and the generalization properties of models trained under supervision.
% Indeed, the loss function must align with one's goals, such that optimization leads to improved model performance.  
For classification problems, Categorical Cross Entropy (CE)~\cite{good1952rational} loss is standard.
For predicted probabilities $p_k$, as-labelled class $\tilde{y}$, Cross Entropy loss is defined as
$\text{CE}(p_k, \tilde{y}) = -\log(p_{\tilde{y}})$.
% \begin{equation}
%     \text{CE}(p_k, \tilde{y}) = -\log(p_{\tilde{y}}).
% \end{equation}
Focal loss (FL)~\cite{lin2017focal} builds upon Cross Entropy loss by placing additional weight on difficult-to-classify samples, through the inclusion of an additional factor, $(1-p_{\tilde{y}})^\gamma$, with weighting parameter $\gamma$ (with a standard setting of $\gamma=2$). 
Focal loss is defined as $\text{FL}(p_k, \tilde{y}) = -(1 - p_{\tilde{y}})^\gamma \log(p_{\tilde{y}})$.
% \begin{equation}
%     \text{FL}(p_k, \tilde{y}) = -(1 - p_{\tilde{y}})^\gamma \log(p_{\tilde{y}}).
% \end{equation}
% Focal loss primarily sees use in classification problems were some classes are more difficult to learn than others, for example in highly imbalanced datasets where some classes have far fewer training examples from which to learn.
Note that if $\gamma=0$, Focal loss reduces simply to Cross Entropy. 
Despite their widespread use, standard loss functions are sensitive to label errors~\cite{ghosh2017robust,patrini2017making,song2022learning} as a result of the strong penalization of confident mispredictions, wherein $p_{\tilde{y}}$ is low for erroneous $\tilde{y}=\hat{y}$.

Robustness to label errors has motivated several alternative loss functions.
Mean Absolute Error (MAE) is theoretically robust to symmetric label corruption~\cite{ghosh2015making,ghosh2017robust}, but has been shown to perform poorly on complicated datasets~\cite{zhang2018generalized}. 
Generalized Cross Entropy (GCE)~\cite{zhang2018generalized} combines Cross Entropy with MAE to obtain the positive characteristics of both losses.
The GCE loss function, officially termed $\mathcal{L}_q$ loss, is defined as the negative Box-Cox transformation\cite{box1964analysis}, 
\begin{equation}
    \text{GCE}(p_k, \tilde{y}) = \mathcal{L}_q(p_k, \tilde{y}) = \frac{(1 - p_{\tilde{y}}^q)}{q},
    \label{eq:gce}
\end{equation}
where $q\in(0, 1]$.
The parameter $q$ controls the transition between these two losses, such that in the limit as $q\to0$, GCE becomes Cross Entropy (via L’Hôpital’s rule), and for $q=1$, GCE becomes MAE. 
The recommended parameter setting is $q=0.7$ for most problems~\cite{zhang2018generalized}.

Active Negative Losses (ANL)~\cite{ye2023active} set the current state-of-the-art and are a class of loss functions that combine Normalized Loss Functions~\cite{ma2020normalized}, $\mathcal{L}_{\mathrm{norm}}$, (originally introduced to be robust to label errors; normalizes by dividing loss for as-labelled class with loss summed over all classes) and Normalized Negative Loss Functions (NNLFs), $\mathcal{L}_{\mathrm{nn}}$, (introduced in~\cite{ye2023active}) through weighting parameters, $a,b>0$,
\begin{equation}
    \mathrm{ANL}(p_k, \tilde{y}) = \alpha \cdot \mathcal{L}_{\mathrm{norm}} + \beta \cdot \mathcal{L}_{\mathrm{nn}}.
    \label{eq:anl}
\end{equation}
Any existing loss function, such as Cross Entropy or Focal loss, can be converted to an ANL through the substitution of normalized and normalized negative forms.
The inclusion of L1 regularization is suggested to avoid overfitting, which commonly occurs for ANL without regularization. 

Other notable robust loss functions include Symmetric Cross Entropy~\cite{wang2019symmetric} (combines Reverse Cross Entropy~\cite{pang2018towards} with CE), PHuber-CE~\cite{menon2020can} (composite loss-based gradient clipping applied to CE), Active Passive Loss (APL)~\cite{ma2020normalized} (designed to maximize $p_{\tilde{y}}$ while minimizing $p_{k}$, $k\ne\tilde{y}$), Asymmetric Loss Functions (ALFs)~\cite{zhou2021asymmetric}, and Curriculum Loss~\cite{lyucurriculum}.
These precursors set the stage for continued research into robust loss design.

\section{Method}
\label{sec:method}
Two novel loss functions are introduced that are designed to be robust to label errors. 
During training on corrupt data, $D_\eta=\{(x_i,\tilde{y_i})\}_{i=1}^N$, mislabelled samples $(x,\hat{y})$ are likely difficult to classify and thus have low predicted probabilities, $p_{\tilde{y}}=p_{\hat{y}}$, for their as-labelled class, $\hat{y}$. 
With CE loss, having low predicted probability, $p_{\tilde{y}}$, means the \textit{gradient} of the loss is \textit{large and negative}, imparting a strong signal to the optimizer causing fitting to these samples.
The proposed loss functions achieve their robustness by reducing the loss for samples likely to be mislabelled, either resulting in a \textit{positive} or \textit{zero} gradient and no longer incorrectly steering the optimizer towards fitting to these erroneous samples.
 
Though the proposed loss functions are the underlying usable contributions to the field, the greater contribution of this article is in the thorough comparative evaluation against other state-of-the-art losses and ablation studies used to provide a deeper understanding of their mechanisms. 
These experiments involve the use of these loss functions within existing label error detection frameworks, Confident Learning (CL) and Area Under the Margin (introduced in \Cref{subsec:label_error_detection}), to correctly identify mislabelled data in artificially corrupted datasets, and compare the effectiveness resulting from the use of these proposed loss functions \vs existing state-of-the-art robust loss functions.
The proposed loss functions are outlined here, whereas the experimental details are outlined in \Cref{sec:experiments}.

\subsection{Blurry Loss}
\textit{Blurry Loss}~\cite{pellegrino2024loss} is closely related to Focal loss~\cite{lin2017focal}, but \textit{de-weights} difficult-to-classify samples, likely to be mislabelled, through a multiplicative factor. 
For predicted probability $p_k$, as-labelled class $\tilde{y}$, and a weighting parameter $\gamma\in\mathbb{R}$, Blurry Loss is defined as
\begin{equation}
    \text{BL}(p_k, \tilde{y}) = -(p_{\tilde{y}})^\gamma \log(p_{\tilde{y}}).
    \label{eq:bl}
\end{equation}
\Cref{fig:BL_loss_graphic} shows the this loss function for several variations of parameter $\gamma$. 

Notice that this function is non-monotonic and features a section of \emph{positive} gradient for $p_{\tilde{y}}<e^{-1/\gamma}$, encouraging training \emph{against} the as-labelled class if $p_{\tilde{y}}$ is low. 
% Derivative is $\frac{d\mathrm{BL}}{dp} = -p^{\gamma-1}(\gamma\log(p)+1)$, and the zero-crossing is at $p=e^{-1/\gamma}$.
At $\gamma=0$, Blurry Loss is equivalent to Cross Entropy Loss.

\subsection{Piecewise-zero Loss}
\begin{figure*}[tb]
    \centering
    \begin{subfigure}[t]{0.48\textwidth}
        \centering
        \includegraphics[width=\textwidth]{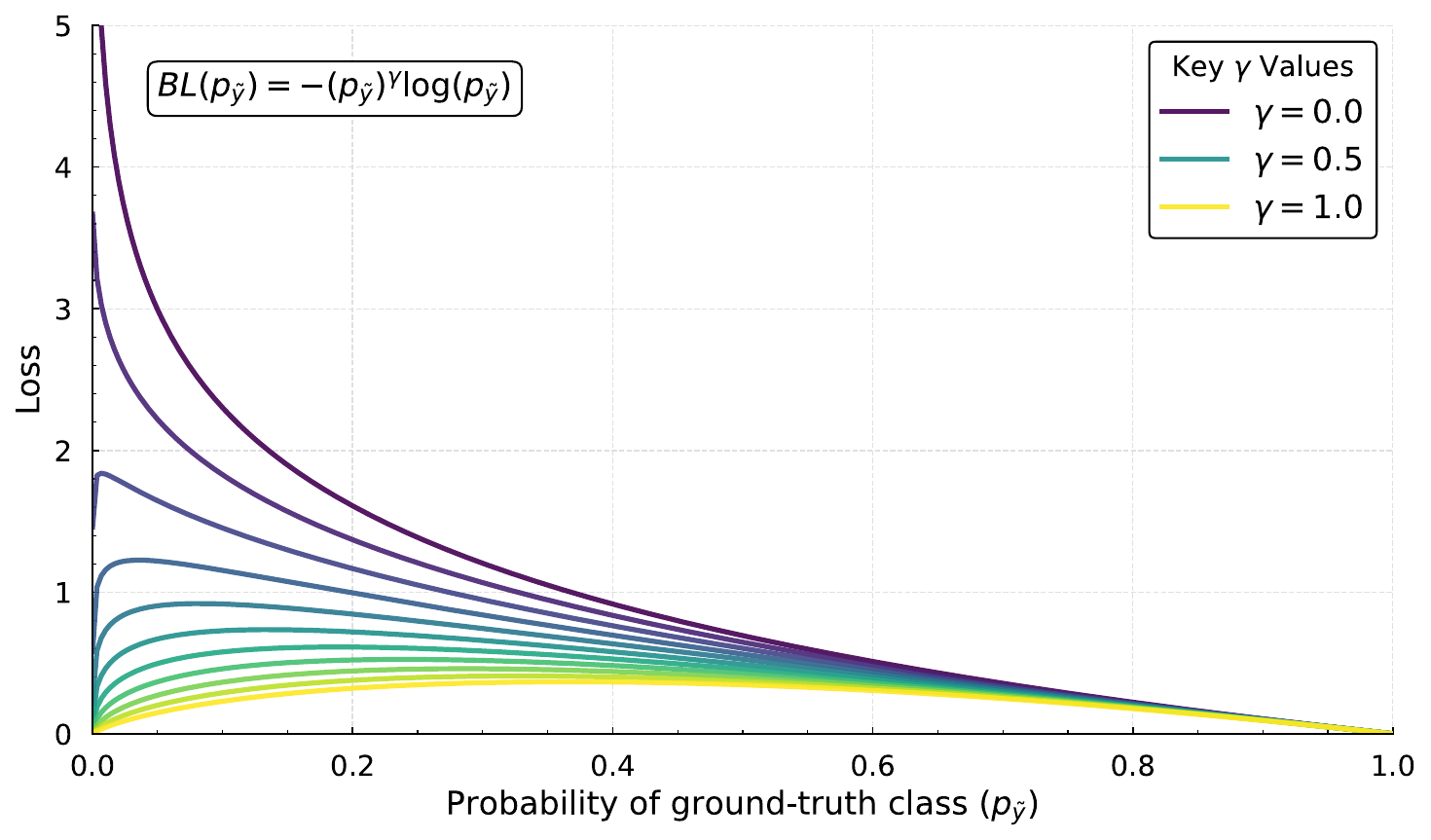}
        \caption{The Blurry Loss function of \Cref{eq:bl}, plotted for a range of parameter $\gamma$, controlling the shape. Note that at $\gamma=0$, this is equivalent to CE.}
        \label{fig:BL_loss_graphic}
    \end{subfigure}%
    ~~~
    \begin{subfigure}[t]{0.48\textwidth}
        \centering
        \includegraphics[width=\textwidth]{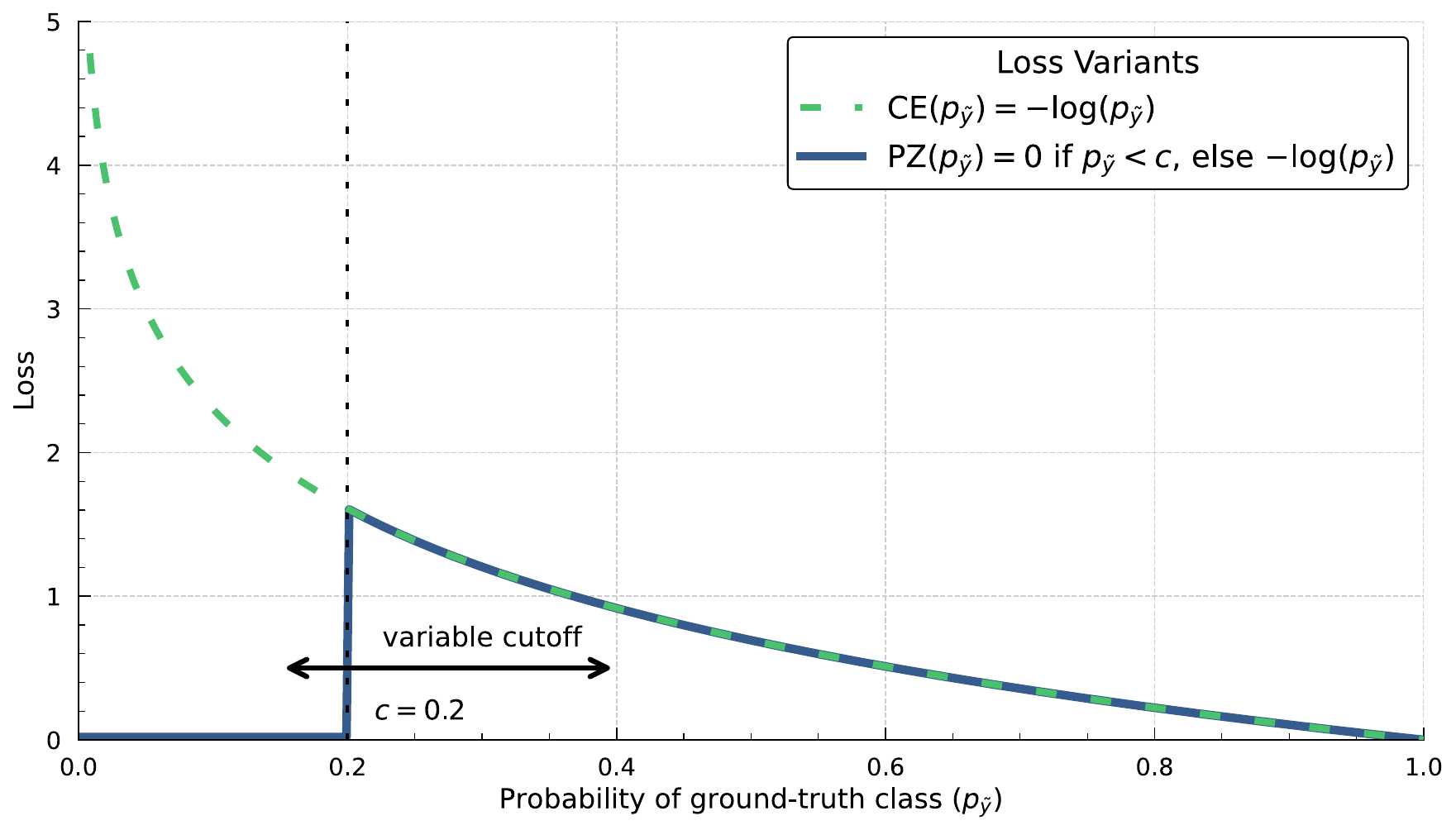}
        \caption{The Piecewise-zero Loss function of \Cref{eq:pz}, illustrating the effect of adjusting the cutoff position, $c$. Note that at $c=0$, this is equivalent to CE.}
        \label{fig:PZ_loss_graphic}
    \end{subfigure}
    \caption{The two proposed loss functions: Blurry Loss (\subref{fig:BL_loss_graphic}) and Piecewise-zero Loss (\subref{fig:PZ_loss_graphic}).}
    \label{fig:loss_graphics}
\end{figure*}

\textit{Piecewise-zero Loss}~\cite{pellegrino2024loss} is a variant of CE loss and is designed to ignore (zero) difficult-to-classify samples, which are likely to be mislabelled.
Examples with predicted probability beneath some cutoff, $p_{\tilde{y}}\leq{c\in[0,1]}$, are assigned a loss of zero through a piecewise definition of the function.
Also note that the \emph{gradient} within the cutoff region is also zero, causing these examples with low $p_{\tilde{y}}$ not to affect the training process (weights are not updated by zero-gradient samples).
Piecewise-zero Loss is defined as
\begin{equation}
    \mathrm{PZ}(p_k, \tilde{y}) = 
    \begin{cases} 
      0 & p_{\tilde{y}}\leq c, \\
      \mathrm{CE}(p_k,\tilde{y}) = -\log(p_{\tilde{y}}) & p_t> c. 
   \end{cases}
   \label{eq:pz}
\end{equation}
\Cref{fig:PZ_loss_graphic} shows this loss function for several variations of the cutoff-position parameter $c$.
At $c=0$, Piecewise-zero Loss is equivalent to Cross Entropy Loss.

\subsection{Loss Scheduling}
DNN model parameters are typically randomly initialized at the start of training.
As a result, the model’s outputs at the early stages of training bear no meaningful relationship to the training labels.
Applying the proposed loss functions immediately --- especially the Piecewise-zero Loss with a high cutoff --- is likely to lead to many correctly labelled samples being de-weighted or ignored, dramatically reducing the ability to train.
Instead, a scheme through which training begins with Cross Entropy loss is proposed, to allow the model to sufficiently generalize prior to the use of these novel loss functions.
A delay parameter, $d\in\mathbb{N}$, determines the epoch at which the training switches to one of the proposed loss functions.
With this scheduling approach, 
% Cross Entropy is used up to epoch $d$, and from epoch $d+1$ onward, training proceeds with the proposed alternative.
for example with PZ loss, the loss, $\mathcal{L}_d$, used is given by
\begin{equation}
    \mathcal{L}_d(p_k, \tilde{y}) = 
    \begin{cases} 
      \mathrm{CE}(p_k,\tilde{y}) & \mathrm{Epoch} \le d, \\
      \mathrm{PZ}(p_k,\tilde{y}) & \mathrm{Epoch} > d. 
   \end{cases}
   \label{eq:delay_loss}
\end{equation}

\section{Experiments}
\label{sec:experiments}

% We will be testing the proposed loss functions for their robustness to label errors during training, as measured by the ability of the label error detection framework to detect label errors when its surrogate model is trained using the proposed loss functions compared to other loss functions. 
% The primary experiment performs this comparison over several datasets and with a range of corruption rates for each.
% An additional variant of this experiment is performed on a pre-cleaned version of one dataset, with only synthetically corrupted labels present, to remove the impacts of existing label errors on the evaluated performance metrics.
% Ablation studies explore the parameters, $\gamma$ and $c$, of the proposed loss functions, as well as the delay, $d$. 
% Further ablation studies examine the case where label errors are non-uniformly distributed, and whether the results of the main experiment hold when using another label error detection framework.
% A final study examines the training dynamics under the proposed loss functions, demonstrating the distributional effects on training examples with and without label errors as evidence that the proposed loss functions operate as desired.

The following paragraphs outline the datasets, method of artificial corruption, loss functions used for comparison, models and training scheme, and metrics used for the experiments.

\paragraph{Datasets and Artificial Corruption}
Synthetic uniform label noise is injected into standard benchmark datasets following the same method used in~\cite{natarajan2013learning,zhang2018generalized,ye2023active,pellegrino2023effects}. 
For a given corruption rate $\eta \in [0, 1]$, a fraction $\eta$ of the training labels are randomly flipped. 
Corruption rates in $\eta \in \{0.1,0.2,0.3,0.4\}$ are tested.
Datasets include MNIST~\cite{deng2012mnist}, Fashion-MNIST~\cite{xiao2017fashion}, CIFAR-10, and CIFAR-100~\cite{krizhevsky2009learning}, under the assumption that their existing label error rates are negligible compared to the induced corruption, $\eta$.

To assess robustness under realistic, non-uniform noise, the CIFAR-10N and CIFAR-100N~\cite{wei2022learning} datasets are employed, which are versions of the CIFAR-10 and CIFAR-100 datasets, re-annotated using Amazon Mechanical Turk. 
By design, the re-annotations do \textit{not} correct labels but rather induce abundant, real-world, non-uniform human annotation errors. 
The original CIFAR labels are considered ground truth in the CIFAR-N datasets.
For CIFAR-10N, three re-annotations are given, whereas for CIFAR-100, only one is given (with $\eta=0.4020$).
With CIFAR-10N, we use the `Aggregate' version (labels used are the majority vote of the three re-annotations; $\eta=0.0903$) and the `Worst' version (re-annotations \textit{not} matching ground truth are always used when available; $\eta=0.4021$). 

\paragraph{Baseline Loss Functions}
Categorical Cross-Entropy~\cite{good1952rational} and Focal Loss~\cite{lin2017focal} are the primary baseline loss functions, as these serve as the \textit{de facto} gold standard for most supervised classification problems. Focal Loss is used with $\gamma = 2$ as recommended in~\cite{lin2017focal}. 
To assess robustness under label errors, we also include state-of-the-art robust loss functions, Generalized Cross-Entropy~\cite{zhang2018generalized} (GCE) loss, defined in \Cref{eq:gce}, Active Negative Cross-Entropy~\cite{ye2023active} (ANL-CE) and Active Negative Focal Loss~\cite{ye2023active} (ANL-FL), defined in \Cref{eq:anl}. 

Consistent with the original formulation, the GCE hyperparameter \textit{q} is fixed at 0.7 as is recommended in~\cite{zhang2018generalized}. 
For the ANL variants, the dataset-specific hyperparameters $\alpha$, $\beta$, and $\delta$, as well as the focal-loss scaling factor $\gamma$ for ANL-FL, are set exactly as prescribed in~\cite{ye2023active}. 
However, since the Fashion-MNIST dataset was not used in~\cite{ye2023active}, no parameter specification was provided and thus the MNIST parameters of~\cite{ye2023active} are used for the Fashion-MNIST experiments here. 
Each loss function is evaluated using these recommended settings to ensure a fair comparison. 

\paragraph{Models and Hyperparameters}
For MNIST and Fashion-MNIST, we adopt a minimal convolutional neural network consisting of two convolutional layers followed by two fully connected layers interleaved with dropout. 
Under a ten-epoch training schedule, this architecture reliably exceeds 99\% test accuracy on clean data (near state-of-the-art)~\cite{kadam2020cnn}.
For CIFAR-10, CIFAR-100, and their noisy variants (CIFAR-10N/CIFAR-100N), we fine-tune an ImageNet-pretrained ResNet-34~\cite{he2016deep} gently modified for lower resolution images.
After training for ten epochs, the model attains 93.5\% test accuracy on CIFAR-10 and 74.5\% on CIFAR-100. Additional details on models and optimizers are given in \Cref{subsec:appendix_exp_details}.

To explore the parameter-sensitivity of our proposed losses, we perform grid sweeps over the Blurry Loss exponent, $\gamma$, and the Piecewise-Zero Loss cutoff, $c$, thereby characterizing their influence on label error detection efficacy and allowing optimal values to be identified for each problem. As demonstrated in ablation \Cref{fig:delay_study} of \Cref{subsec:appendix_additional_studies}, introducing a one-epoch delay yields the best performance for PZ loss; accordingly, we use $d=1$ for all subsequent PZ loss experiments.

\paragraph{Label Error Detection and Performance Metrics} 
To evaluate each loss function’s efficacy for the detection of label errors, we apply the Confident Learning (CL)~\cite{northcutt2021confident} and AUM~\cite{pleiss2020identifying} frameworks, both described in \Cref{subsec:label_error_detection}.  
The CL pruning method used in all cases is ``both'', in which detected samples must both be unlikely to be their as-labelled class and likely to be of another different class.
Five cross-validation folds are used with 80\% train / 20\% predict splits.
The use of AUM serves as an ablation study of label error detection, to demonstrate the efficacy of the proposed loss functions across multiple detection frameworks. 
Following the methodology of \cite{pleiss2020identifying}, experiments on CIFAR-100 are replicated, using a ResNet-32~\cite{he2016deep} model trained for 150 epochs with a learning rate of 0.1 and weight decay of $10^{-4}$; however, with Cross-Entropy (CE)~\cite{good1952rational} loss replaced with Blurry and Piecewise Zero Loss for a direct comparison.

Performance is primarily quantified using the F1 score~\cite{van1979information} --- a gold-standard metric defined as the harmonic mean of precision and recall. 
The F1 score penalizes situations in which either precision or recall fall, which is not always reflective of one's goals when attempting to identify label errors.
% (\eg if corruption rates are quite low, a condition of high imbalance between corrupt and clean samples, slightly sacrificing precision if it means greater recall may be worthwhile, since relatively few detections will be made).
To address this limitation with low corruption rates, we also report the Balanced Accuracy score~\cite{5597285}, defined as the average of sensitivity (true positive rate) and specificity (true negative rate), which provides a more robust measure of detector performance under class imbalance.

\paragraph{Compute Resource Requirements}
All experiments were conducted on a single NVIDIA Tesla V100-SXM2 GPU (16 GB VRAM). Reproducing the main CL label error detection results requires about 6~GB of GPU memory and takes roughly 10 minutes per loss-function and hyperparameter combination. AUM experiments demand 7~GB of memory and approximately 2.5 hours per loss-function and hyperparameter test. Further details are given in \Cref{subsec:appendix_exp_details}.

\section{Results}
\label{sec:results}

Results for the main experiment (\Cref{subsec:main_findings}) and additional studies (\Cref{subsec:ablation_studies}) are presented here. All tables are formatted such that best scores are \underline{\textbf{bolded and underlined}}, and second best are simply \textbf{bolded}.

\subsection{Main Findings}
\label{subsec:main_findings}

The proposed loss functions are compared against existing standard and robust loss functions, including CE, FL, GCE, ANL-CE, and ANL-PZ. 
This comparison is performed on several datasets and corruption rates, 
%ranging from $\eta=0.1$ to $0.4$, 
as described in \Cref{sec:experiments}. 
%Note that lower rates, towards 0.1 are more realistic than 0.4, where 40\% of the training data are corrupt. 
Results from 20 random trials are summarized in \Cref{tab:chart_df_F1}.
The proposed loss functions are evaluated over a range of their parameters, $\gamma$ and $c$; however, only the best performing cases are shown for brevity. 
In nearly all cases, the use of the proposed BL and PZ losses result in better performance than with the baseline loss functions. 

\input{figures/main_study_F1} 

At low corruption rates ($\eta=0.1$), FL occasionally performs well as consequence of it causing CL to detect more conservatively, favouring precision over recall compared to the proposed loss functions. 
At these low rates, the detection problem is highly imbalanced:  high recall may be achieved while making few correct detections, but precision is dramatically reduced with relatively few false detections. 
In a context where detected examples are reviewed by experts, prioritizing recall over precision can be advantageous, as catching more errors generally outweighs the cost of incorrectly detecting a few more examples. 
% Clearly this trade-off depends on one's own goals and abilities; however, 
Precision and recall are examined for BL and PZ, and compared to FL in \Cref{fig:lowCR_Prec_Rec_F1} of \Cref{subsec:appendix_additional_studies}, where it is found that by tuning the parameters of the proposed loss functions, recall is increased significantly, but at the cost of some precision.  
To capture a more reasonable objective at lower corruption rates, Balanced Accuracy (mean of sensitivity and specificity) is reported in \Cref{tab:chart_df_BA}, in which 
the proposed BL and PZ perform strongly throughout.

\input{figures/main_study_BA}

% Precision and recall (the two harmonic components of the F1 score) are explored for BL and PZ, and compared to FL in \Cref{fig:lowCR_Prec_Rec_F1} of \Cref{subsec:appendix_additional_studies}. 
% It is found that by tuning the parameters of the proposed loss functions, recall is increased significantly, but at the cost of some precision. 
% Because F1 is a harmonic mean, the small decrease in precision dominates the larger increase in recall, and the score is decreased.
% Again, this does not necessarily indicate less desirable behaviour at lower corruption rates, since having improved recall may be worthwhile at the expense of having relatively few false-positives. 

In the main experiment of \Cref{tab:chart_df_F1,tab:chart_df_BA}, the corrupted datasets include pre-existing, unaccounted for label errors.
While the usage of several datasets demonstrated the broad applicability and efficacy of the proposed loss functions, an additional experiment on a dataset without pre-existing label errors would enable a more precise measure of label error detection performance. 
To achieve this, a version of CIFAR-100 with test samples having human-verified label errors removed~\cite{northcutt2021pervasive} is used. 
The cleaned test set is artificially corrupted in the same manner as in the main experiment, and the Confident Learning error detection framework is once again applied; however, rather than using a k-fold approach over the entire dataset, model training is performed strictly on the non-cleaned training set, and evaluation is performed on the cleaned (and artificially corrupted) test set. 
Results averaged over three seeds are shown in \Cref{tab:clean_CIFAR_results} and form a direct comparison to the CIFAR-100 experiments shown in \Cref{tab:chart_df_F1}.
The performance of all loss functions is improved, especially at lower corruption rates (improvement of nearly 0.1 in F1 score at $\eta=0.1$).
This improvement at $\eta=0.1$ matches expectations given that the proportion of pre-existing label errors is more significant at lower artificial corruption rates.
The most dramatic improvement is seen for PZ loss for all $\eta$, which performs best in all cases.

\input{figures/pruned_CIFAR100}

\subsection{Ablation Studies}
\label{subsec:ablation_studies}

\paragraph{Non-uniform Corruption}
The main experiment included only \textit{uniform} artificial corruption.
To better understand the performance of the proposed loss functions under the presence of \textit{non-uniform} label errors, a study involving the CIFAR-10N and CIFAR-100N~\cite{wei2022learning} datasets with real-world human annotation errors is performed. 
% Note that the re-annotations of the CIFAR-N datasets do not correct labels but rather induce real-world human annotation label errors. 
This study follows the same format as the main experiment.
Results averaged over five random seeds are shown in \Cref{tab:CIFAR_N_f1_comparison_std} against baseline loss functions, and for three variations of the CIFAR-N datasets, as described in \Cref{sec:experiments}. 
Rows are ordered with increasing corruption rate. 
Notice that the proposed loss functions are best (and second best) performing in all cases, and the optimal parameter setting tends to increase with corruption rate, congruent with the results of ablation \Cref{fig:parameters_vs_cr} in \Cref{subsec:appendix_additional_studies}.

\input{figures/CIFAR-N_chart_horizontal_std}

\paragraph{Alternative Label Error Detection Framework}
The main experiment used Confident Learning as the label error detection framework.
To demonstrate the effectiveness and versatility of the proposed loss functions when used with other frameworks, an ablation study using the AUM~\cite{pleiss2020identifying} framework is performed on the CIFAR-100 dataset. 
Resulting label error detection F1 scores averaged over three random seeds are summarized in \Cref{tab:aum_summary} and are directly comparable to the CIFAR-100 results of \Cref{tab:chart_df_F1}. 
The label error detection performance exceeds that of Confident Learning, and the proposed loss functions improve upon the baseline.

\input{figures/AUM_CIFAR-100}

\paragraph{Training Dynamics}

\begin{figure}[tb]
    \centering
    \begin{subfigure}[b]{0.32\textwidth}
        \centering
        \includegraphics[width=\linewidth]{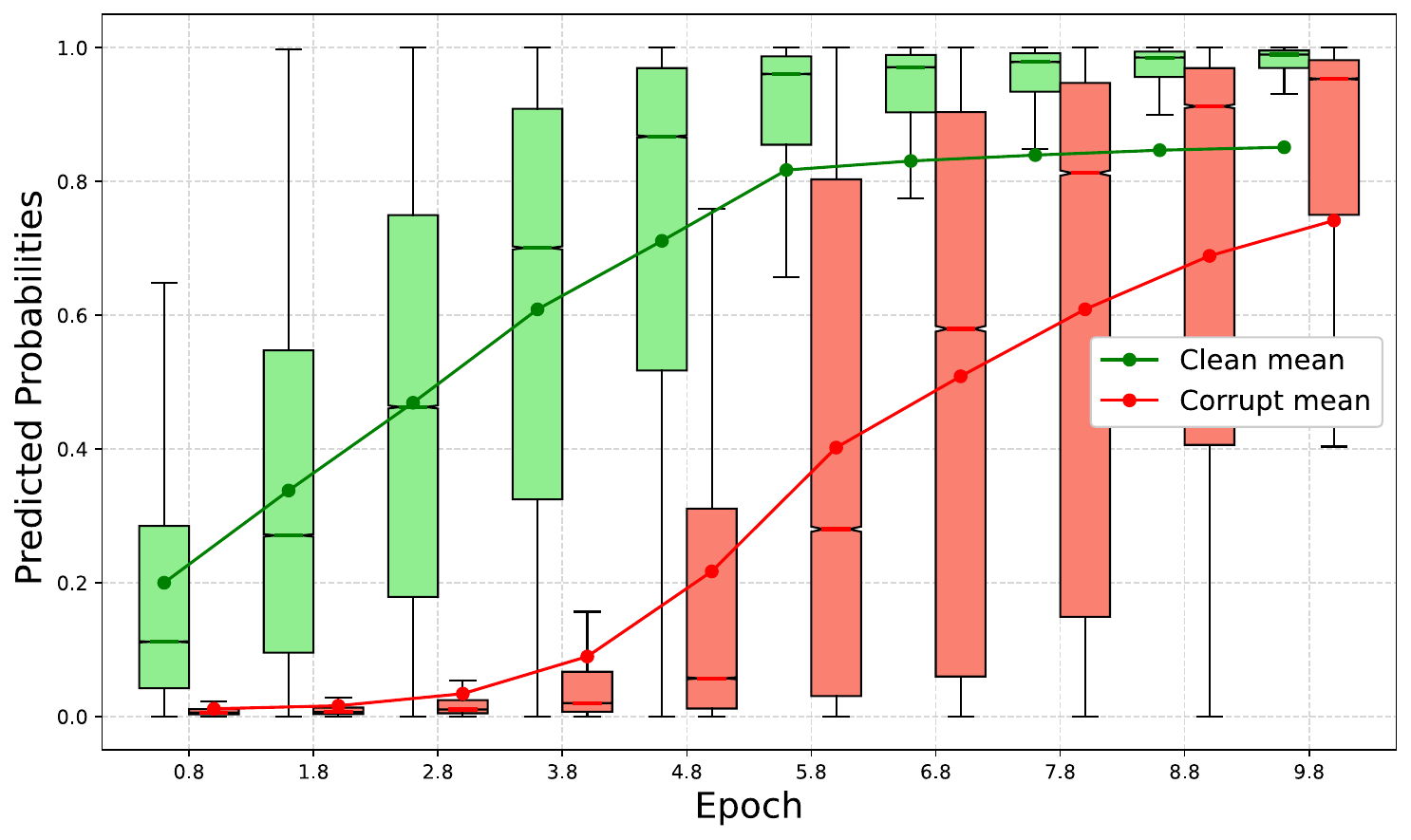}
    \end{subfigure}
    \hfill
    \begin{subfigure}[b]{0.32\textwidth}
        \centering
        \includegraphics[width=\linewidth]{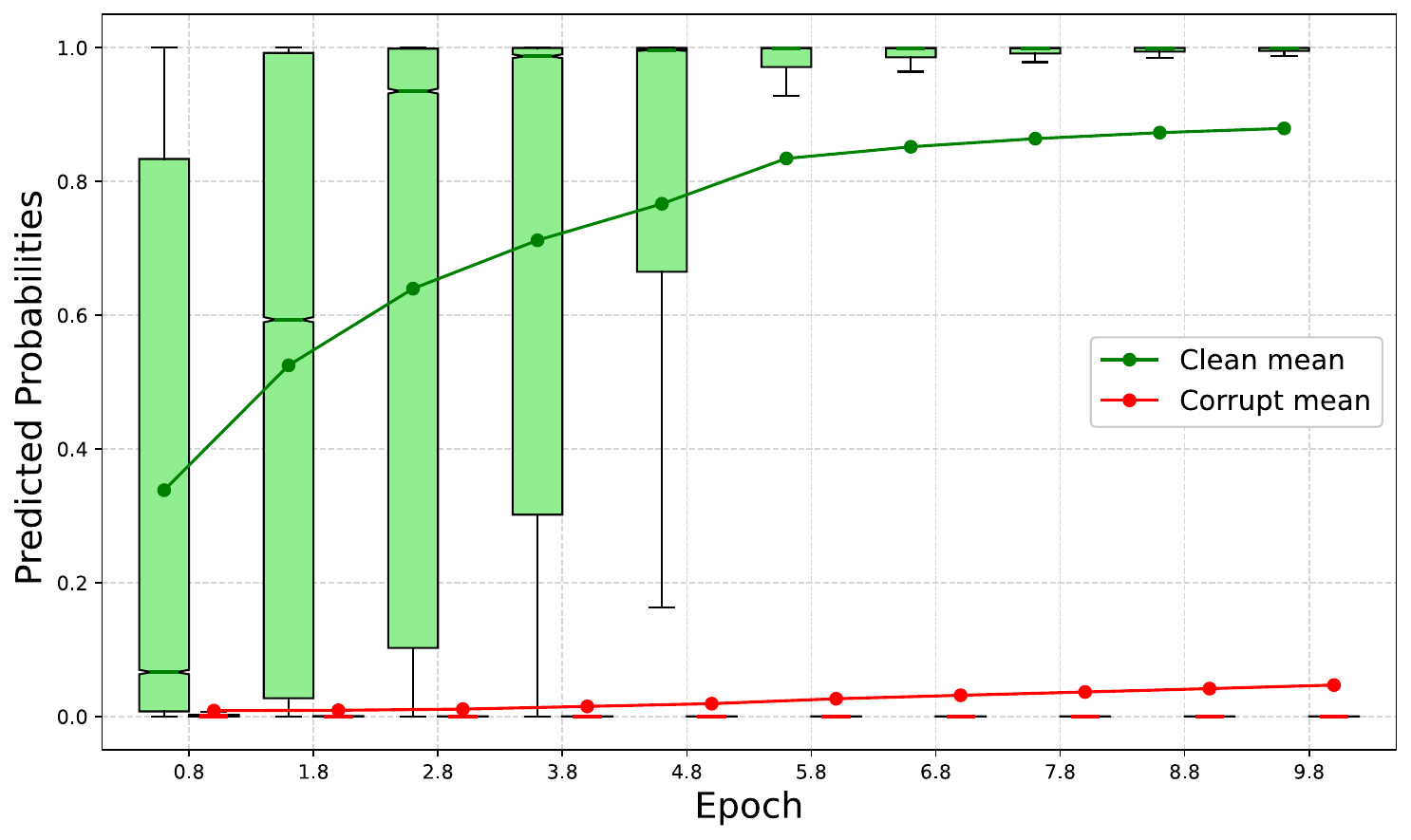}
    \end{subfigure}
    \hfill
    \begin{subfigure}[b]{0.32\textwidth}
        \centering
        \includegraphics[width=\linewidth]{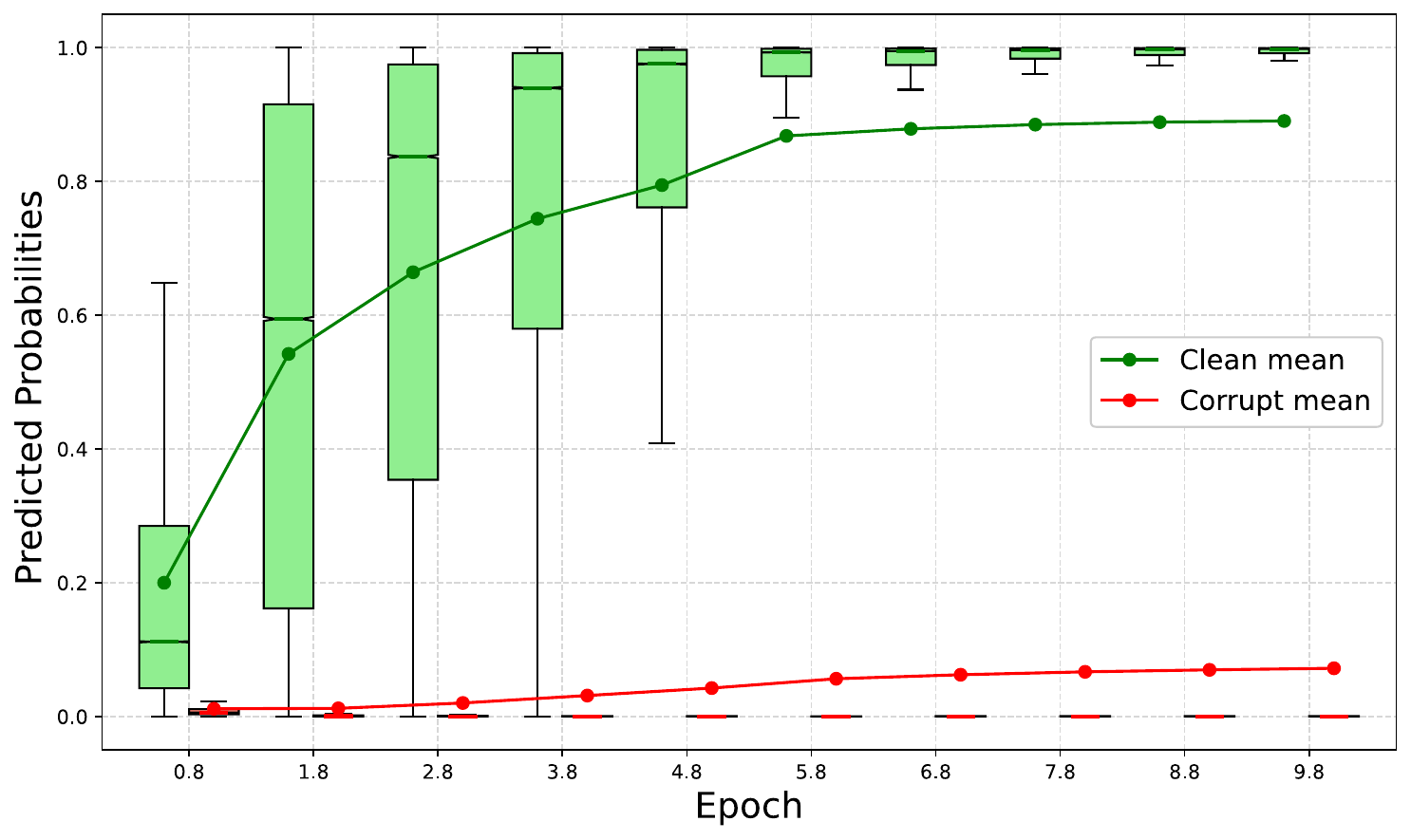}
    \end{subfigure}
    
    \begin{subfigure}[b]{0.32\textwidth}
        \centering
        \includegraphics[width=\linewidth]{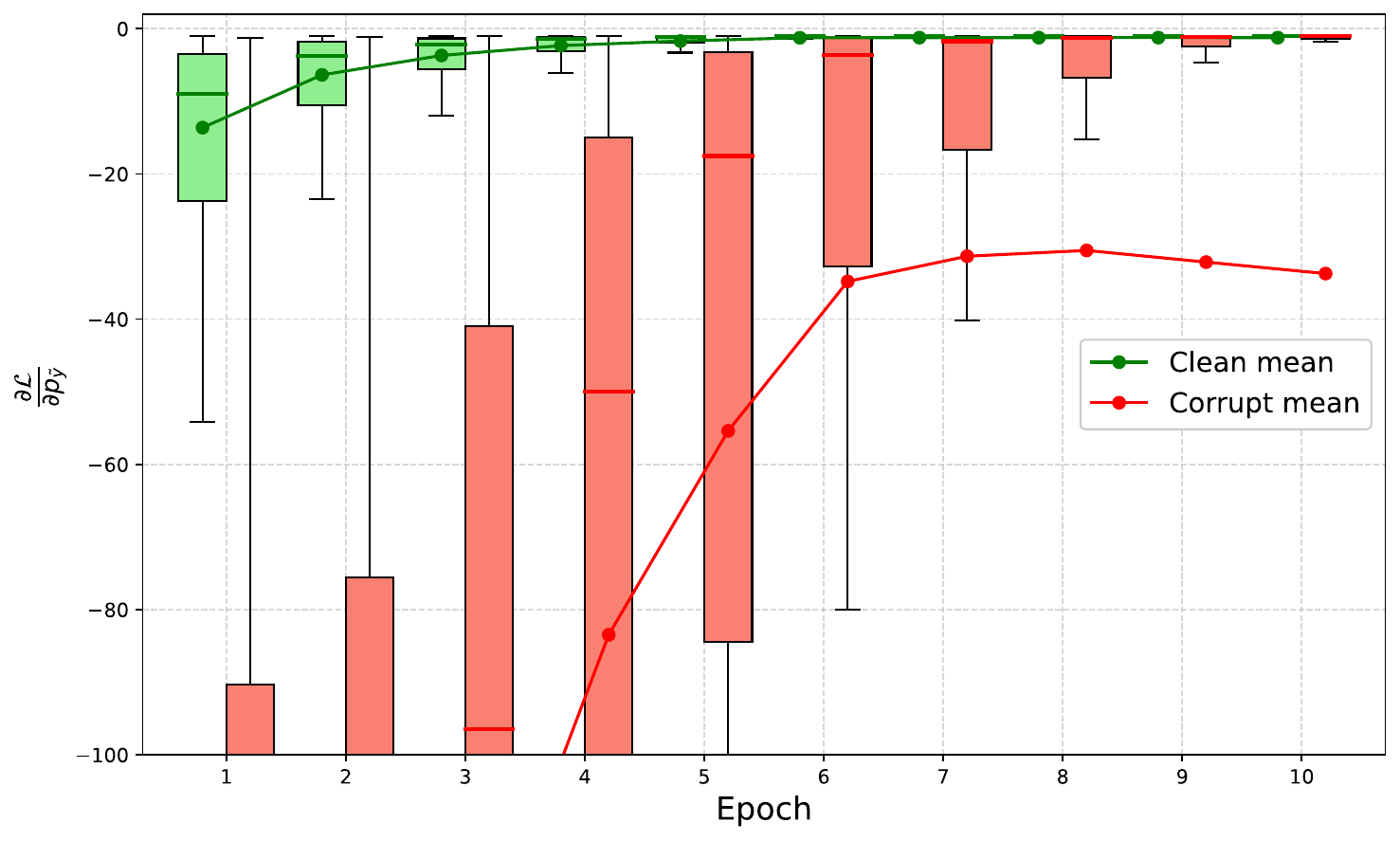}
        \caption{CE}
        \label{fig:gradients_graph_CE}
    \end{subfigure}
    \hfill
    \begin{subfigure}[b]{0.32\textwidth}
        \centering
        \includegraphics[width=\linewidth]{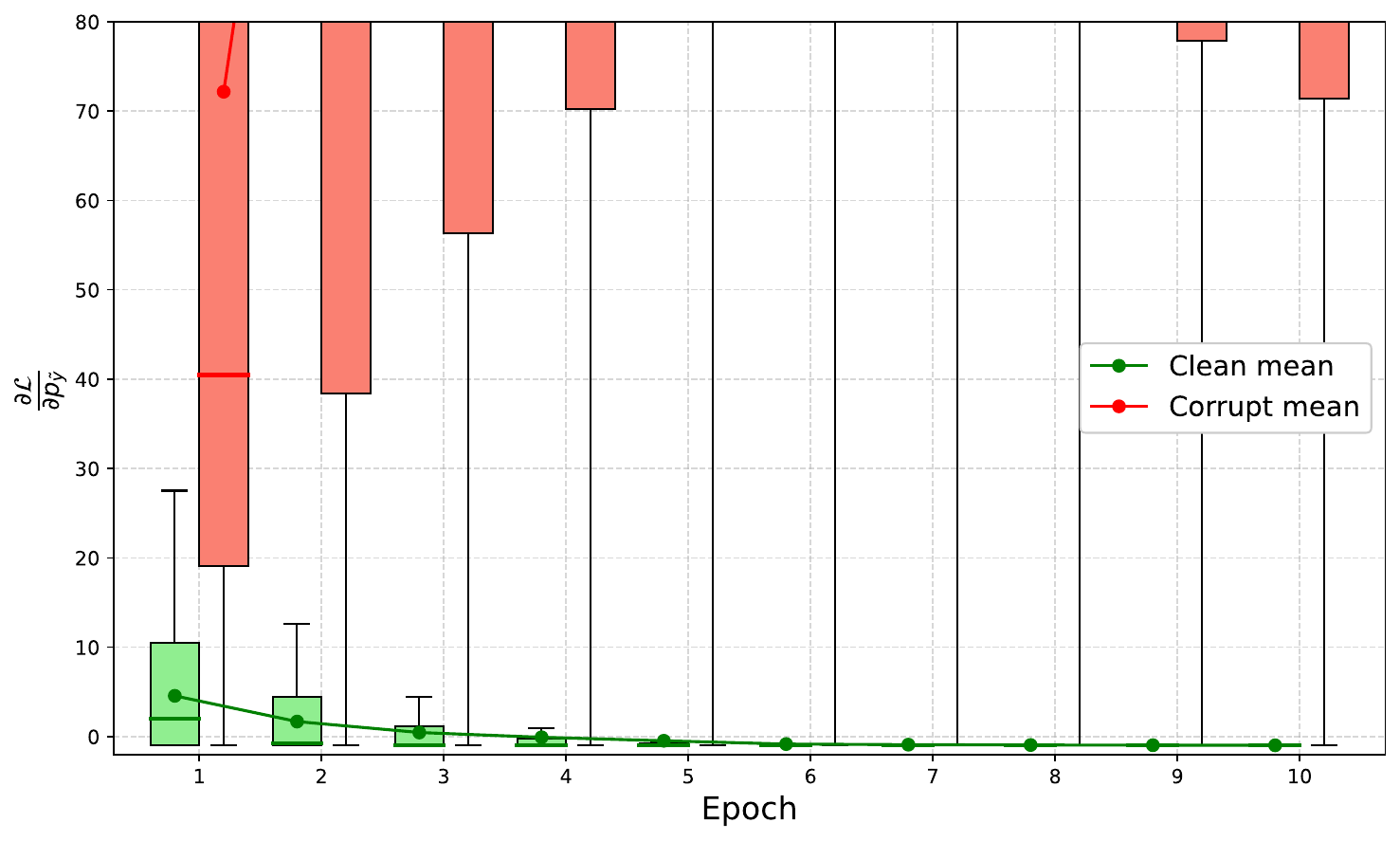}
        \caption{BL}
        \label{fig:gradients_graph_BL}
    \end{subfigure}
    \hfill
    \begin{subfigure}[b]{0.32\textwidth}
        \centering
        \includegraphics[width=\linewidth]{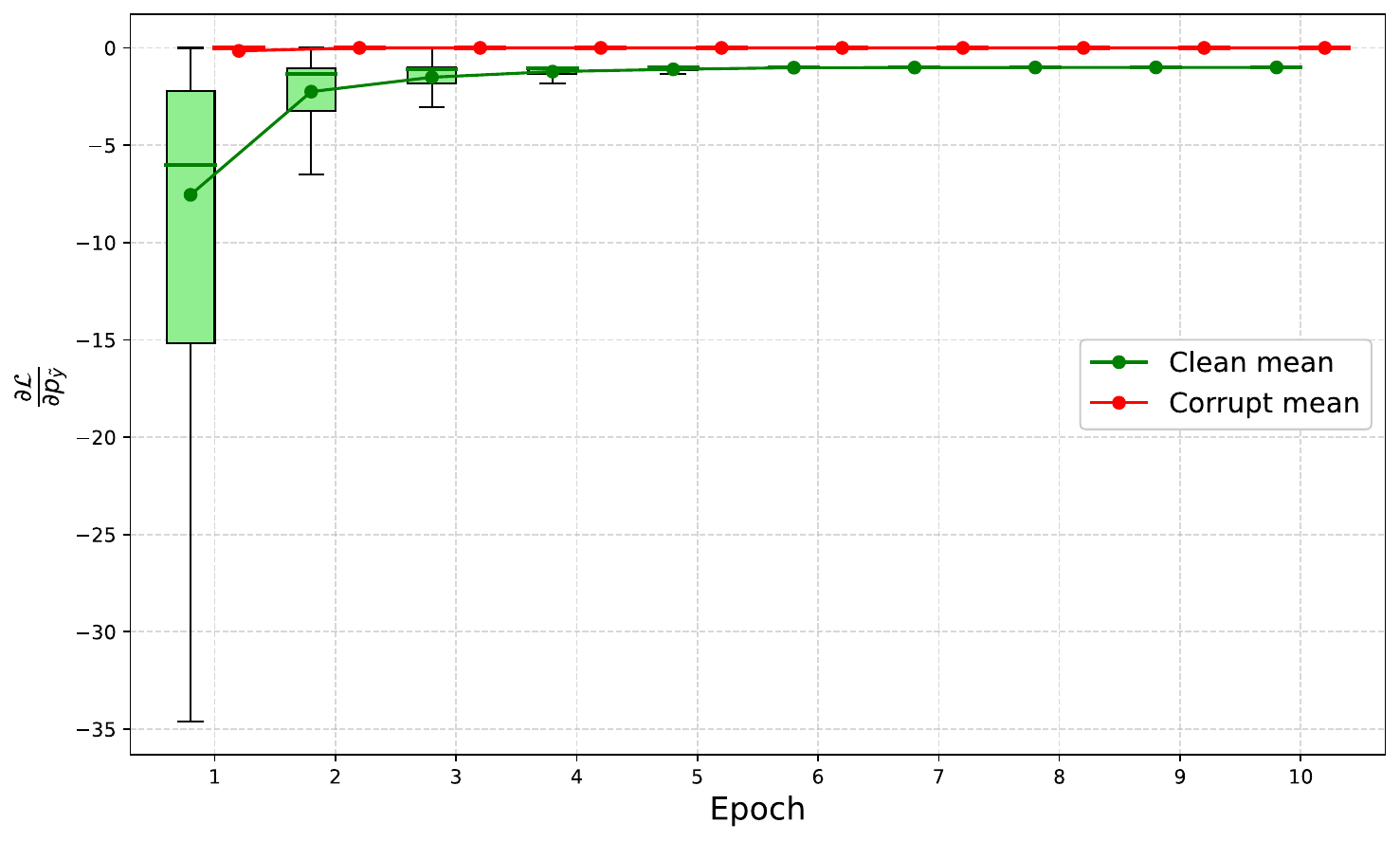}
        \caption{PZ}
        \label{fig:gradients_graph_PZ}
    \end{subfigure}
    \caption{Predicted probability (top) and Gradient (bottom) distributions per epoch, for CIFAR-100 at $\eta=0.4$. With all loss functions, the predicted probabilities of corrupt data (red) are less than those of clean data (green). With CE, the gradients of corrupt data are large and negative, providing a strong signal to the optimizer to fit to these corrupt data, whereas with BL, gradients of corrupt data are large and \textit{positive}, steering away from these corrupt data, and with PZ, gradients of corrupt data are nearly all at zero, imparting no impact on training. A version of this figure without truncated vertical limits can be found in \Cref{fig:gradients_Full} of \Cref{subsec:appendix_expanded}.} 
    \label{fig:gradient_dist}
\end{figure}

To verify that the proposed loss functions affect gradients during training as they are designed to do, the gradient of loss, $\partial \mathcal{L} / \partial p_{\tilde{y}}$, is monitored during training for corrupted and clean samples. 
Box and whisker plots in \Cref{fig:gradient_dist} indicate the distribution of predicted probabilities (top) and gradients (bottom) for corrupt (red) and clean (green) data at each epoch during training. 
CIFAR-100 with $\eta=0.4$ is used for this study.
With all loss functions, the predicted probabilities of corrupt data are less than those of clean data; however, the difference is far greater with the proposed losses. 
With CE, the gradients of corrupt data are large and negative, providing a strong signal to the optimizer to fit to these corrupt data, whereas with BL, gradients of corrupt data are mainly large but \textit{positive}, steering the training away from these corrupt data, and with PZ, gradients of corrupt data are nearly all at zero, imparting no impact on training.
Notice that with BL, the gradients for clean data during the first epochs are mostly \textit{positive}, but quickly settle to a moderate \textit{negative} value, associated with predicted probabilities close to 1.0 (as can be seen in the top row).

% \begin{figure}[tb]
%     \centering
%     \begin{subfigure}[b]{0.32\textwidth}
%         \centering
%         \includegraphics[width=\linewidth]{figures/pred_probs_CE_CIFAR100_04.pdf}
%         \caption{CE}
%         \label{fig:pred_probs_CE}
%     \end{subfigure}
%     \hfill
%     \begin{subfigure}[b]{0.32\textwidth}
%         \centering
%         \includegraphics[width=\linewidth]{figures/pred_probs_BL_CIFAR100_04.pdf}
%         \caption{BL}
%         \label{fig:pred_probs_BL}
%     \end{subfigure}
%     \hfill
%     \begin{subfigure}[b]{0.32\textwidth}
%         \centering
%         \includegraphics[width=\linewidth]{figures/pred_probs_PZ_CIFAR100_04.pdf}
%         \caption{PZ}
%         \label{fig:pred_probs_PZ}
%     \end{subfigure}
%     \caption{Predicted probability distributions, CIFAR100 CR0.4.}
%     \label{fig:pred_prob_dist}
% \end{figure}

\paragraph{Further Studies}
Additional investigations and ablation studies are provided in \Cref{subsec:appendix_additional_studies}, including 
a study observing the precision-recall trade-off and their impact on F1 score, in \Cref{fig:lowCR_Prec_Rec_F1}, 
a study investigating the impacts of the proposed loss function parameters, in \Cref{fig:parameters_vs_cr},
an ablation study on the effects of the delay parameter, $d$, in \Cref{fig:delay_study}, and
an additional study observing differences in distributions of the AUM statistic for corrupt and clean samples, in \Cref{fig:training_dyn_AUM}.

\section{Conclusion and Limitations}
\label{sec:conclusion}
This work provides an in-depth empirical evaluation of two recently proposed loss functions, Blurry Loss and Piecewise-zero Loss, for improving model robustness and the downstream detection of label errors. 
By de-weighting or ignoring samples likely to be mislabelled, these losses enhance the performance of label error detection frameworks across a variety of artificial corruption settings.
Extensive experiments and ablation studies demonstrate their broad applicability and competitive performance relative to existing robust loss functions. 
One limitation of the experiments is that parameter sweeps were not performed for the existing robust loss functions; however, their recommended settings for each dataset were used wherever possible.
Similarly, there is no universally optimal parameter setting for the proposed loss functions.
These findings support the use of such loss functions as practical tools for improving data quality and model reliability in supervised learning scenarios fraught with label errors.

% Include after review, to avoid revealing identifying information during review process
\section*{Acknowledgments}

This research was enabled in part by support provided by Calcul Québec (\small\url{calculquebec.ca}) and the Digital Research Alliance of Canada (\small\url{alliancecan.ca}).

We acknowledge the support of the Government of Canada’s New Frontiers in Research Fund (NFRF), [NFRFT-2020-00073], and the support of the Natural Sciences and Engineering Research Council of Canada (NSERC) via \mbox{NSERC-CGS D}.

Nous remercions le Fonds Nouvelles Frontières en Recherche du gouvernement du Canada de son soutien (FNFR), [FNFRT-2020-00073], et le soutien du Conseil de Recherches en Sciences Naturelles et en Génie du Canada (CRSNG), CRSNG-BESC D.
\begin{figure}[h]
    \centering
    \includegraphics[width=\columnwidth]{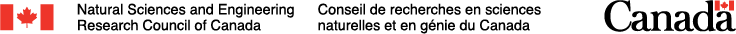}
\end{figure}

% \section*{References}
\bibliographystyle{plainnat} %ieeetr also mostly works, but doesnt' seem to work for \citet{}
\bibliography{references.bib}

%%%%%%%%%%%%%%%%%%%%%%%%%%%%%%%%%%%%%%%%%%%%%%%%%%%%%%%%%%%%

\appendix

\section{Technical Appendices and Supplementary Material}
\label{sec:appendix}

\subsection{Additional Experimental Details}
\label{subsec:appendix_exp_details}

Further information building upon the experimental details of \Cref{sec:experiments} is given here.

\paragraph{Models and Optimizers}
For MNIST and Fashion-MNIST, we adopt a minimal convolutional neural network consisting of two convolutional layers followed by two fully connected layers interleaved with dropout, totalling approximately 1.2~million trainable parameters. 
Under a ten-epoch training schedule, this architecture reliably exceeds 99\% test accuracy (near state-of-the-art)~\cite{kadam2020cnn}.

For CIFAR-10, CIFAR-100, and their noisy variants (CIFAR-10N/CIFAR-100N), we fine-tune an ImageNet-pretrained ResNet-34~\cite{he2016deep} modified only by replacing its initial $7\times7$, stride-2 convolution with a $3\times3$, stride-1 convolution and removing the following max-pool layer, preserving spatial resolution on $32\times32$ inputs.
We train for ten epochs under identical conditions using the Adam optimizer~\cite{kingma2014adam} with an initial learning rate of $1\times10^{-4}$; a scheduler reduces the learning rate by a factor of 0.1 at epoch~5 for CIFAR-10 (factor 0.2 for CIFAR-100). Under this setup, the model attains 93.5\% test accuracy on CIFAR-10 and 74.5\% on CIFAR-100.

All experiments utilize the Adam optimizer, leveraging its adaptive learning capabilities without requiring additional tuning.

\paragraph{Compute Resource Requirements}
All experiments were conducted on a compute cluster using an NVIDIA Tesla V100-SXM2-16GB GPU. The main Confident Learning experiments required 6~GB of memory, and each loss function and parameter combination required on average 10 minutes (5 minutes for MNIST and Fashion-MNIST, 15 minutes for CIFAR10 and CIFAR100). Experiments on the cleaned CIFAR-100 test set followed the same process, requiring only 3~GB memory. When additionally storing the per-epoch gradient information, for MNIST 3~GB of memory and 1 hour was required, and 6~GB of memory and 4 hours for CIFAR100. The CIFAR-N experiments required 15 minutes and 6~GB of memory. The AUM experiments required 7~GB of memory as well as 2.5 hours for each loss function and parameter test conducted. Our complete set of experiments required approximately 1800 hours of compute, with an additional $\sim400$ hours for preliminary experimentation.

The Confident Learning experiments were conducted using the Python library \texttt{cleanlab} v2.5.0, from \url{https://github.com/cleanlab/cleanlab}, as introduced in \citet{northcutt2021confident} under an AGPL-3.0 license. The AUM experiments were conducted using the python library \texttt{aum} v1.0.2, from \url{https://github.com/asappresearch/aum} as introduced in \citet{pleiss2020identifying} under an MIT license.

\subsection{Additional Experiments and Studies}
\label{subsec:appendix_additional_studies}

\paragraph{Precision-Recall Trade-off in F1 Score}

\begin{figure}[tb]
  \centering
  \begin{subfigure}[b]{0.48\textwidth}
    \centering
    \includegraphics[width=\textwidth]{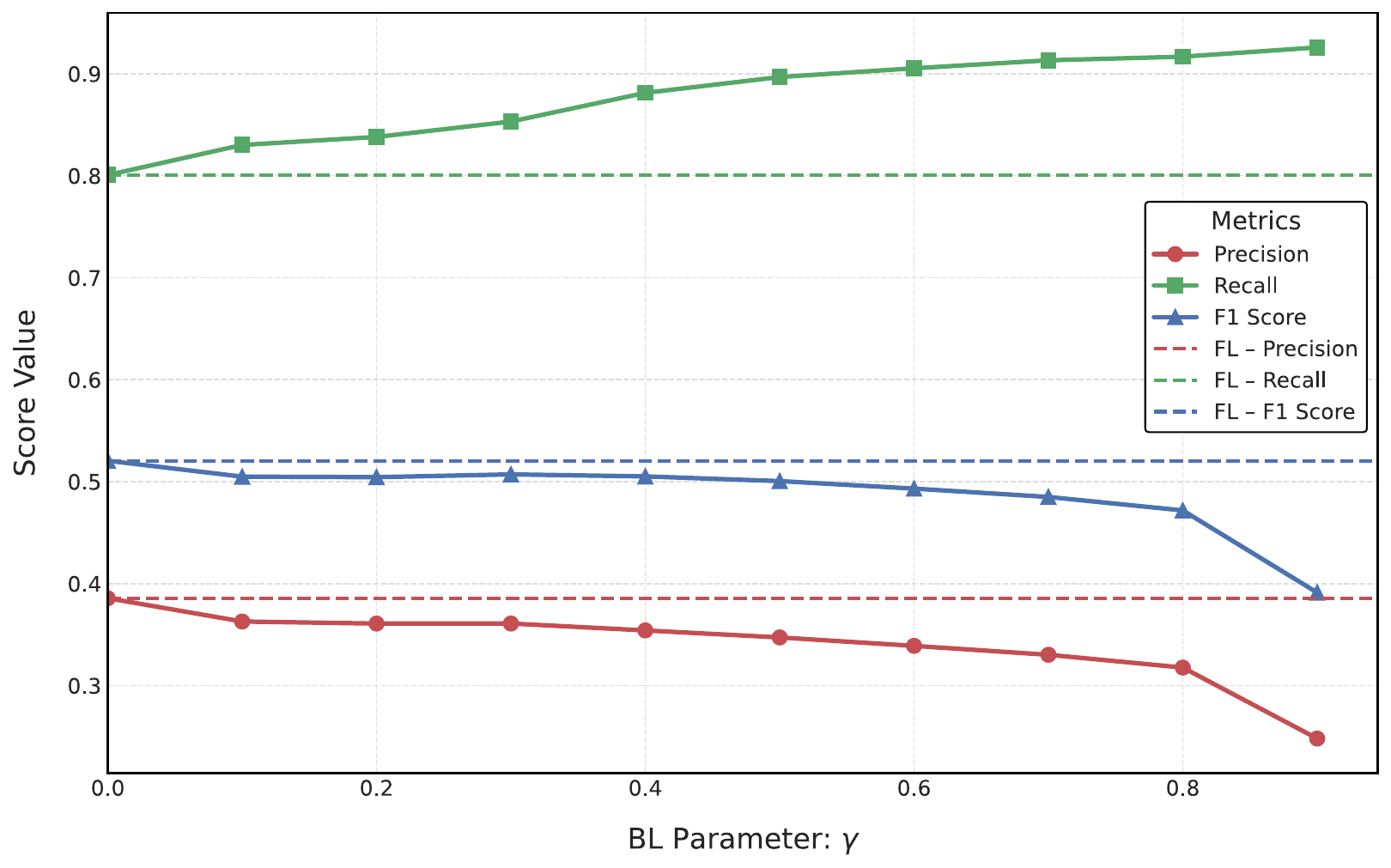}
    \caption{BL \vs FL, CIFAR-100, $\eta=0.1$.}
    \label{fig:lowCR_Prec_Rec_F1_BL}
  \end{subfigure}
  \hfill
  \begin{subfigure}[b]{0.48\textwidth}
    \centering
    \includegraphics[width=\textwidth]{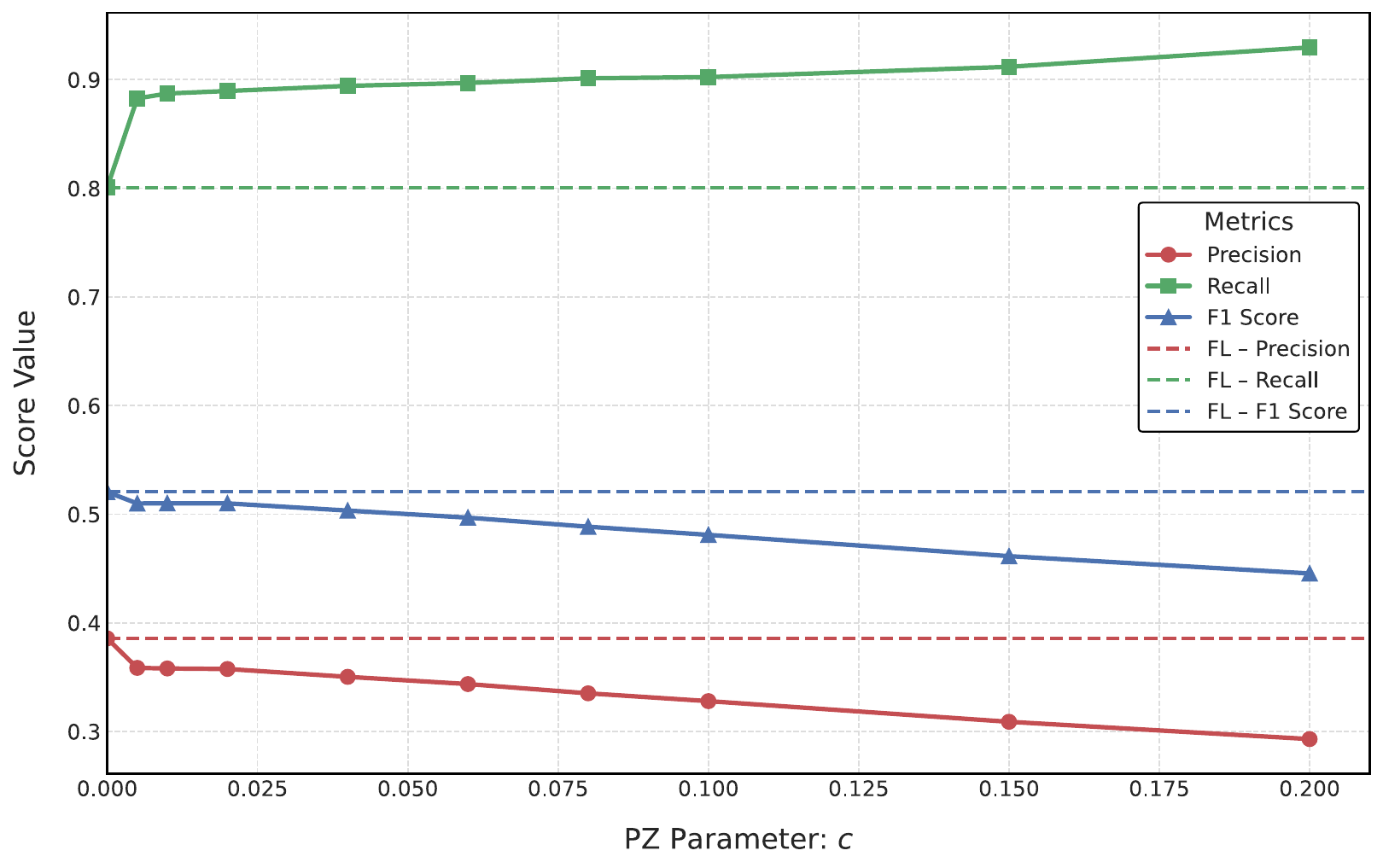}
    \caption{PZ \vs FL, CIFAR-100, $\eta=0.1$.}
    \label{fig:lowCR_Prec_Rec_F1_PZ}
  \end{subfigure}
  \caption{FL (dotted lines) results in better F1 scores than BL (\subref{fig:lowCR_Prec_Rec_F1_BL}) or PZ (\subref{fig:lowCR_Prec_Rec_F1_PZ}) (solid lines) at $\eta=0.1$ on the CIFAR-100 dataset. This is a result of conservative detection, whereby fewer samples are detected and precision is maintained. Through tuning the parameters of the proposed loss functions, performance can be made fairly similar to that of FL, or recall can be increased dramatically while only marginally reducing precision, which may be considered worthwhile at low corruption rates.}
  \label{fig:lowCR_Prec_Rec_F1}
\end{figure}

Precision and recall (the two harmonic components of the F1 score) are explored for BL and PZ, and compared to FL in \Cref{fig:lowCR_Prec_Rec_F1}, for the CIFAR-100 dataset at $\eta=0.1$, a case selected since the F1 score of FL exceeds that of BL and PZ. 
Observe that by tuning the parameters of the proposed loss functions, recall is increased significantly, but at the cost of some precision. 
Because F1 is a harmonic mean, the small decrease in precision dominates the larger increase in recall, and the score is decreased.
Again, this does not necessarily indicate less desirable behaviour at lower corruption rates, since having improved recall may be worthwhile at the expense of having relatively few false-positives. 

\begin{figure}[tb]
\centering
\begin{subfigure}[b]{\textwidth}
   \includegraphics[width=1\linewidth]{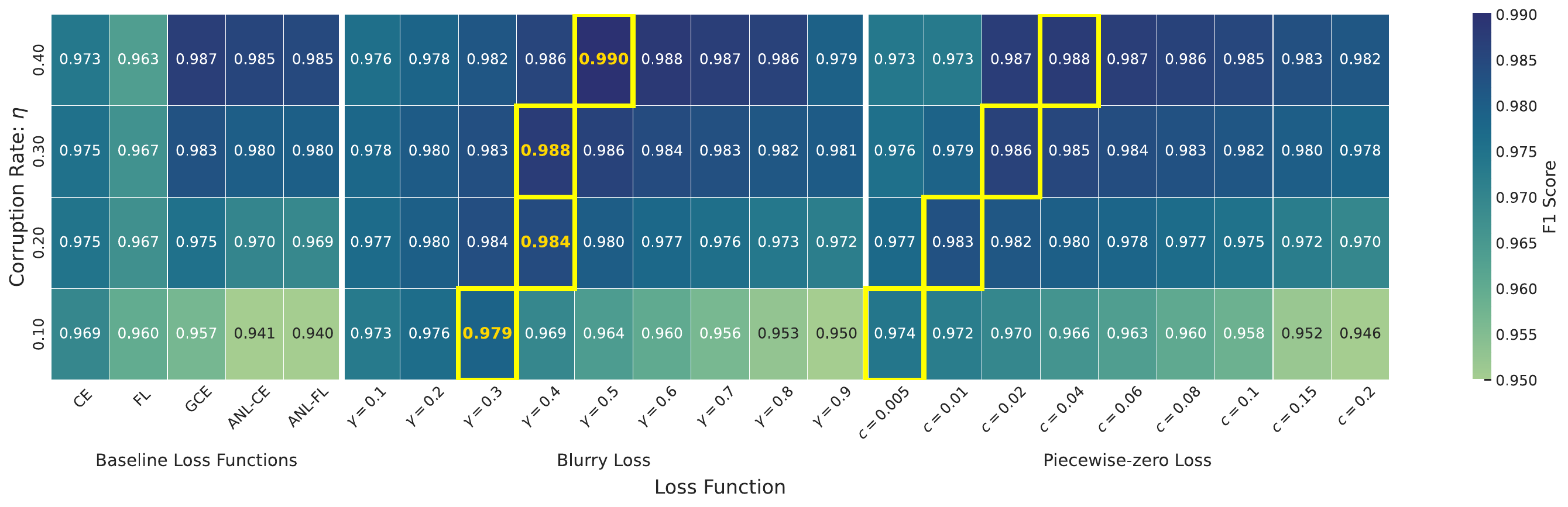}
   \caption{MNIST}
   \label{fig:parameters_vs_cr_MNIST} 
\end{subfigure}

\begin{subfigure}[b]{\textwidth}
   \includegraphics[width=1\linewidth]{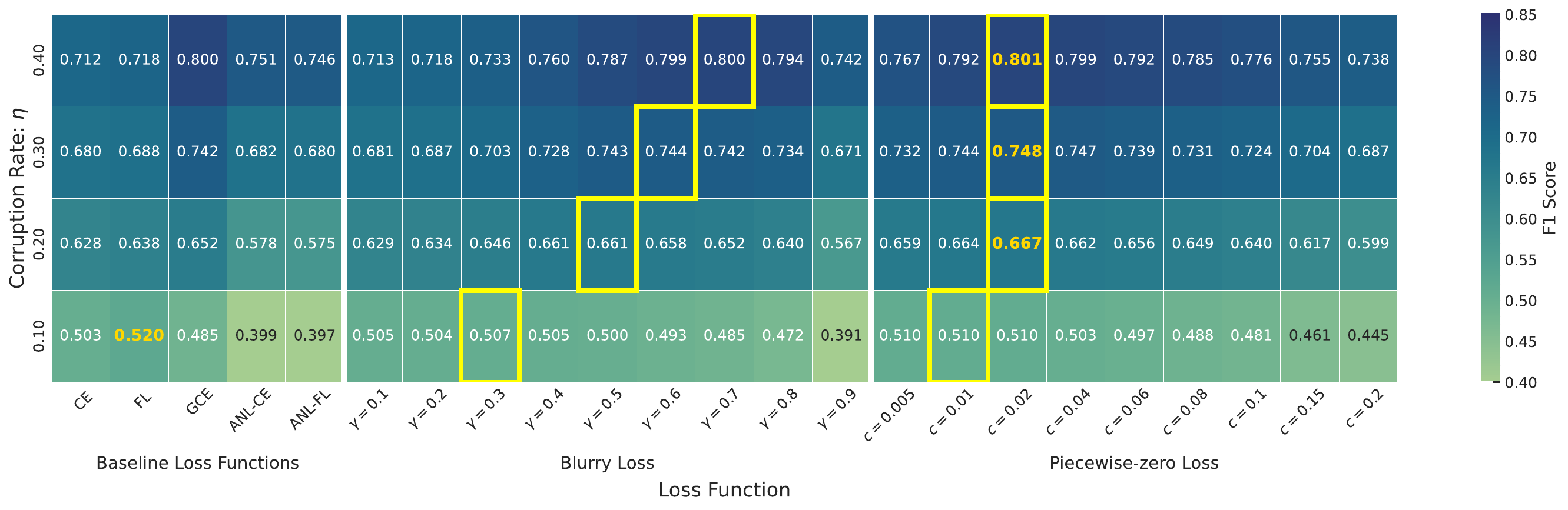}
   \caption{CIFAR-100}
   \label{fig:parameters_vs_cr_CIFAR100}
\end{subfigure}

\caption{Heatmap showing variation in F1 score as loss function parameters are varied, for MNIST (\subref{fig:parameters_vs_cr_MNIST}) and CIFAR-100 (\subref{fig:parameters_vs_cr_CIFAR100}). In the left blocks, separated by a white line, results for baseline loss functions are shown. Two separate blocks are shown for the proposed loss functions, plotted over a range in their parameters. Best results are indicated with bolded yellow text. To illustrate the trend in optimal parameter settings for the proposed loss functions, yellow boxes mark the best performing parameter setting for each row. Note that delay, $d$, is set to 0 for BL and 1 for PZ loss.}
\label{fig:parameters_vs_cr}
\end{figure}

\paragraph{Loss Function Parameters}
To better understand the parameters of the proposed loss functions, an ablation study on these parameters is performed using a similar experimental setup as the main experiment. 
For each dataset and artificial corruption rate, the optimal parameter setting is identified. 
For these experiments, the delay, $d$, is set to 0 for BL and 1 for PZ loss.
Results for MNIST and CIFAR-100 datasets are shown in \Cref{fig:parameters_vs_cr} as a heatmap of F1 score, along with a comparison to the baseline loss functions. 
For each corruption rate (row), the best performing configuration is indicated with bolded yellow text. 
To illustrate the trend in optimal parameter settings for the proposed loss functions, yellow boxes mark the best performing settings.
Observe that optimal settings of both $\gamma$ and cutoff, $c$, increase with increased corruption rate. 
This trend is consistent across datasets; however, more complicated / difficult datasets, such as CIFAR-100 relative to MNIST, seem also to demand greater settings of $\gamma$ (but not $c$).
Note that larger settings for $\gamma$ and $c$ correspond to the proposed loss functions `ignoring' a greater range of predicted probabilities. 

\paragraph{Delay}
Delay, $d$, is introduced to allow the model to generalize prior to training with the proposed loss functions. 
This is especially important for PZ loss, which, at the initial stages of training, outright `ignores' a large proportion of the dataset.
To better understand how best to set the delay, an ablation study on this parameter is performed. 
Again, label errors are detected and performance is measured with F1 score, while $d$ is varied. 
This study was performed across all the main experiment datasets and corruption rates, with fairly consistent results. 

\begin{figure}[tb]
  \centering
  \begin{subfigure}[b]{0.24\textwidth}
    \centering
    \includegraphics[width=\linewidth]{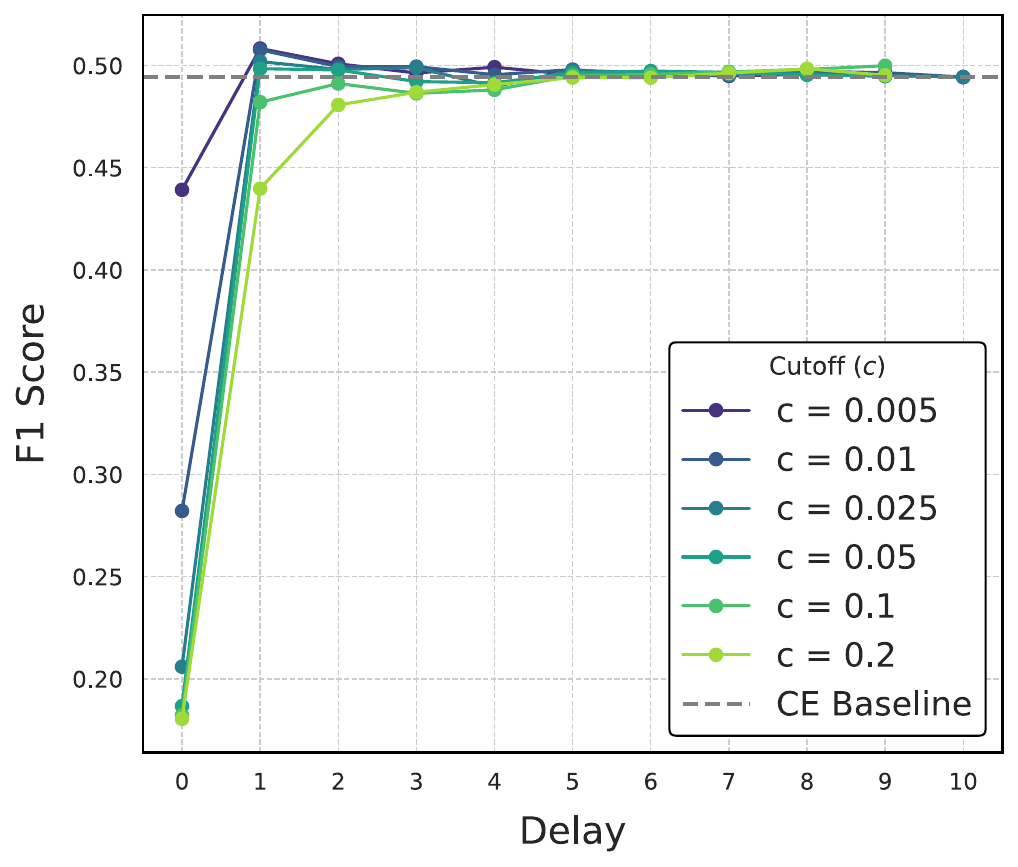}
    \caption{$\eta=0.1$}
    \label{fig:pz_delay_cr1}
  \end{subfigure}
  \hfill
  \begin{subfigure}[b]{0.24\textwidth}
    \centering
    \includegraphics[width=\linewidth]{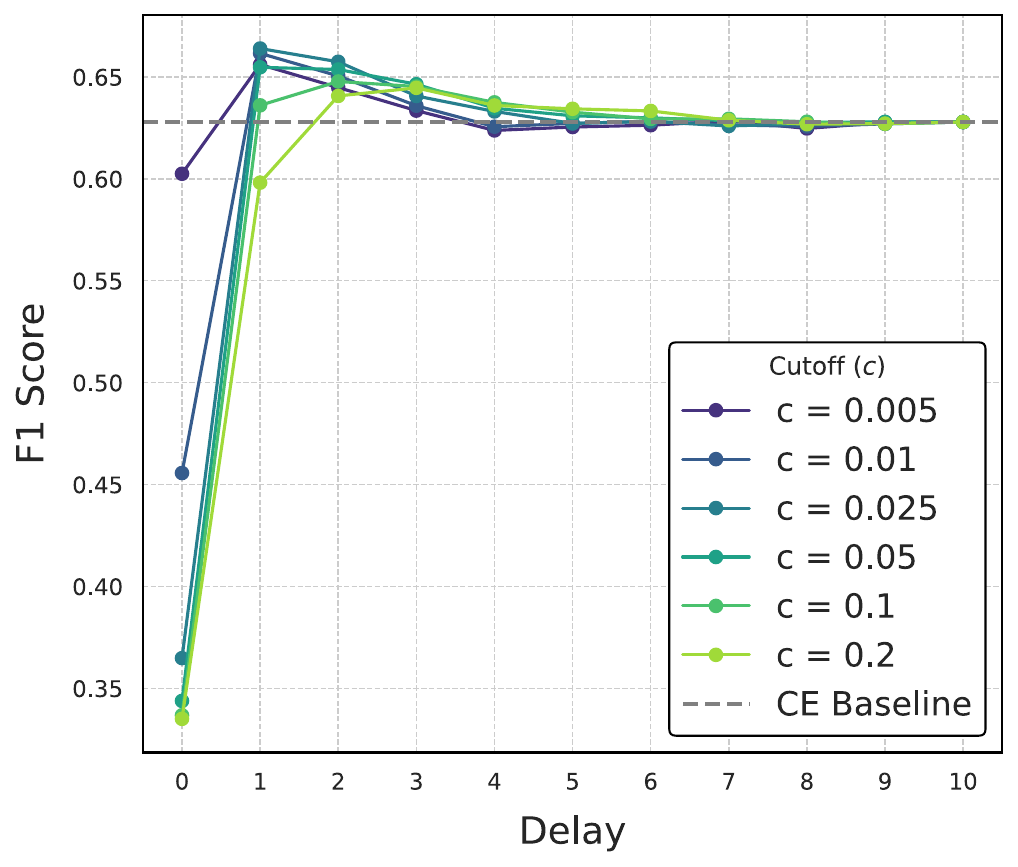}
    \caption{$\eta=0.2$}
    \label{fig:pz_delay_cr2}
  \end{subfigure}
  \hfill
  \begin{subfigure}[b]{0.24\textwidth}
    \centering
    \includegraphics[width=\linewidth]{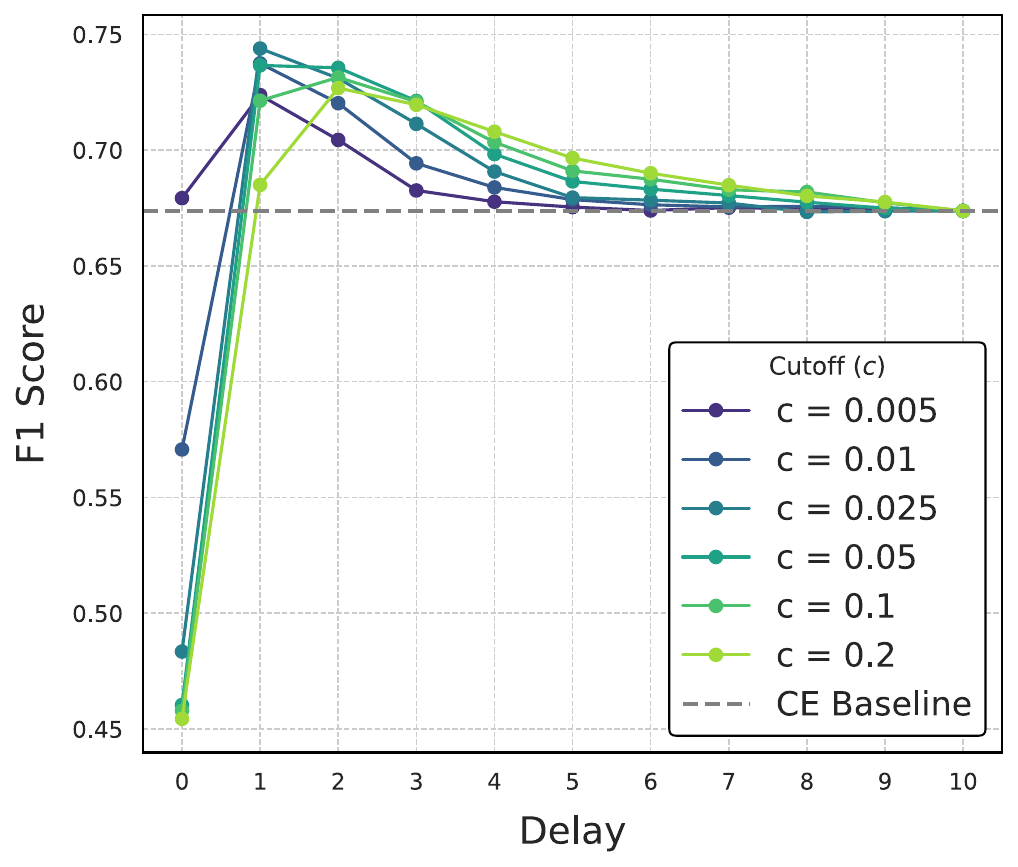}
    \caption{$\eta=0.3$}
    \label{fig:pz_delay_cr3}
  \end{subfigure}
  \hfill
  \begin{subfigure}[b]{0.24\textwidth}
    \centering
    \includegraphics[width=\linewidth]{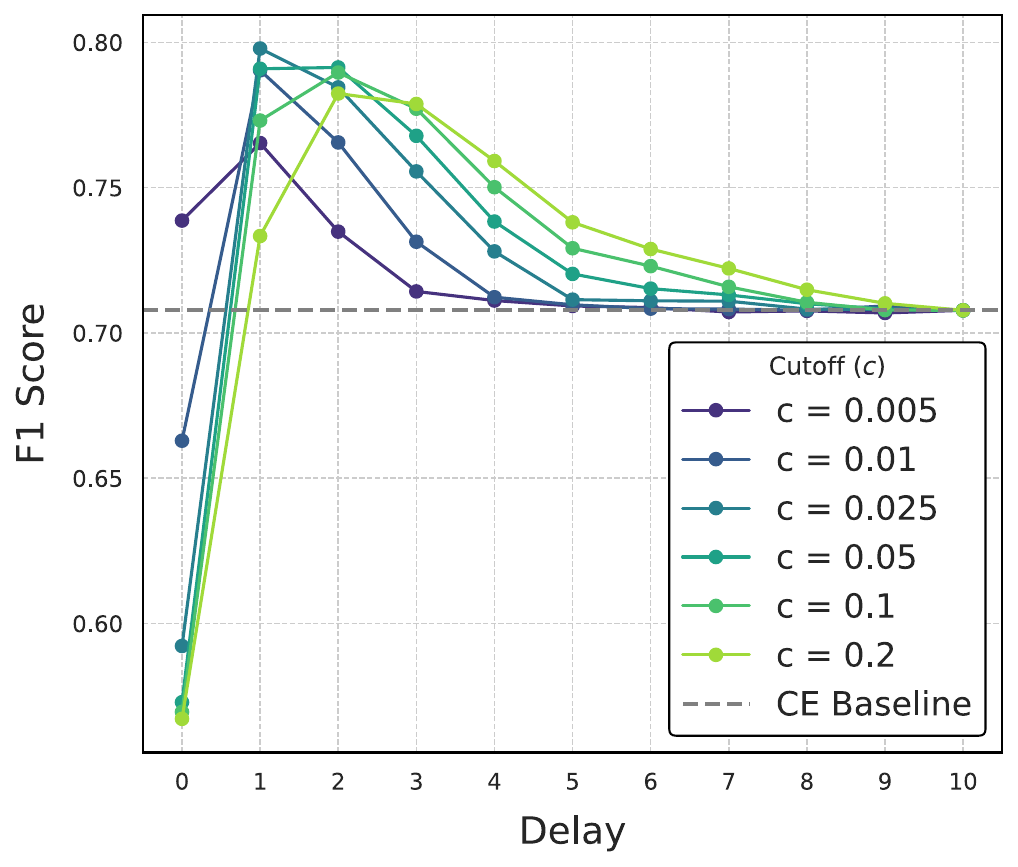}
    \caption{$\eta=0.4$}
    \label{fig:pz_delay_cr4}
  \end{subfigure}
  \caption{Detection F1 score \vs Delay, $d$, for PZ loss on CIFAR-100 at $\eta$ ranging from $0.1$ to $0.4$ in panels (\subref{fig:pz_delay_cr1}) to (\subref{fig:pz_delay_cr4}). Observe that the best performance comes at $d=1$ in all cases, for an appropriate setting of the cutoff, $c$.}
  \label{fig:delay_study}
\end{figure}

Results for the most complex dataset, CIFAR-100, are given in \Cref{fig:delay_study}, for corruption rates ranging from $\eta=0.1$ to $0.4$.
Trends for various cutoff, $c$, settings are plotted.
Observe that there is a large jump in F1 from $d=0$ to $d=1$ (for all $\eta$), indicating that some delay is highly beneficial and allowing the model to somewhat generalize early in training is necessary for good application of PZ loss. 
Additionally, the best performance is achieved for $d=1$ (for an appropriate setting of $c$), and F1 scores tend to towards the CE baseline (poorer performance) as delay is further increased.
This indicates that having a greater number of epochs trained with PZ loss improves label error detection once the model is sufficiently generalized (one epoch with CE), and perhaps even that detrimental fitting to the erroneous data occurs with further training with CE.
Furthermore, notice that the degree of downwards trending relates to the corruption rate, indicating that high corruption rates necessitate a greater need for training with PZ loss to achieve high label error detection performance.

Although not shown here, a similar preliminary study was performed for BL, which indicated that having no delay was always best, and thus delay is suggested only for PZ loss. 
This is likely a result of PZ loss being far more harsh than BL (outright ignoring of data rather than simply having positive gradients for samples with low predicted probabilities).

\paragraph{Training Dynamics}

\begin{figure}[tb]
    \centering
    \begin{subfigure}[b]{0.32\textwidth}
        \centering
        \includegraphics[width=\linewidth]{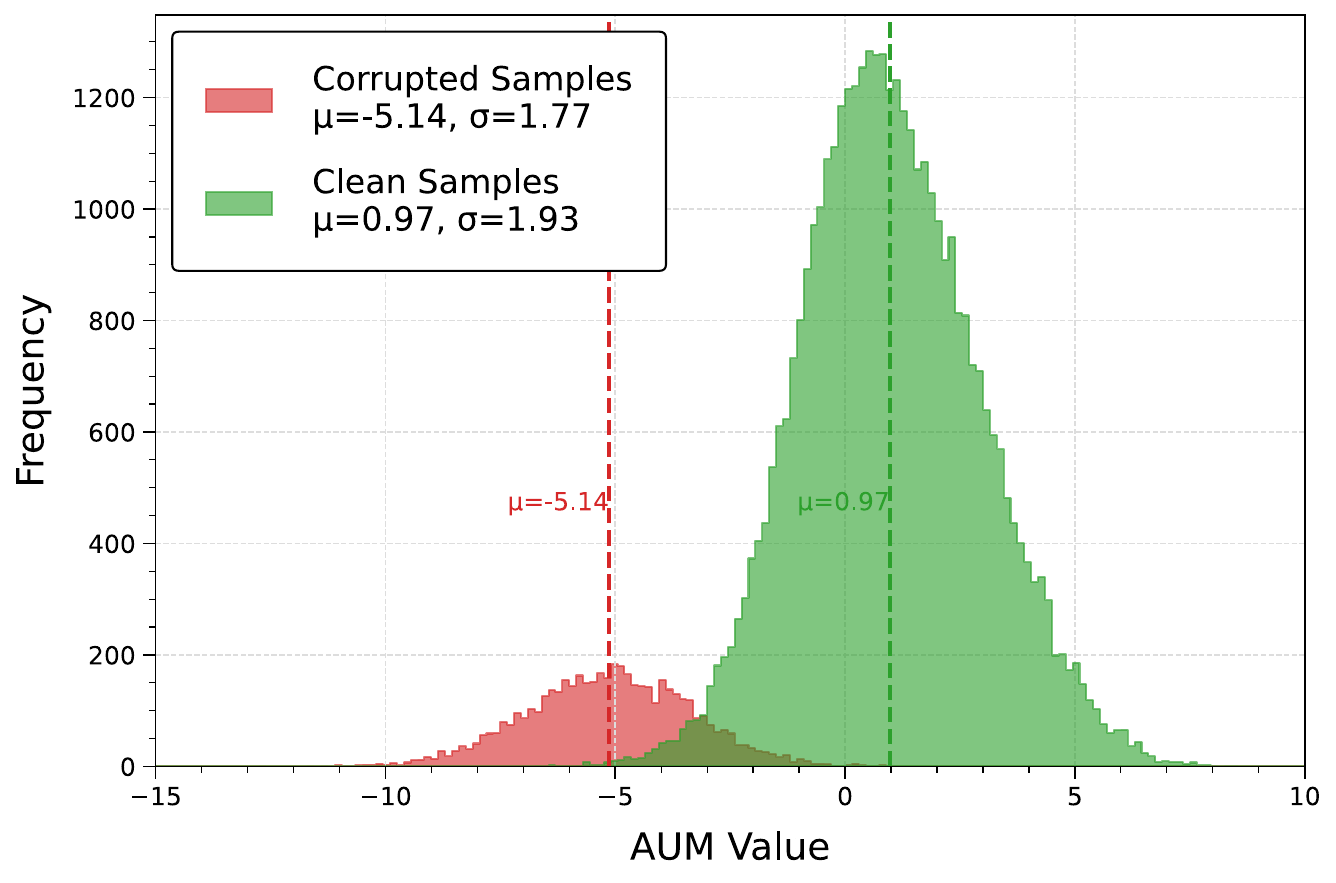}
        \caption{CE}
        \label{fig:aum_graph_CE}
    \end{subfigure}
    \hfill
    \begin{subfigure}[b]{0.32\textwidth}
        \centering
        \includegraphics[width=\linewidth]{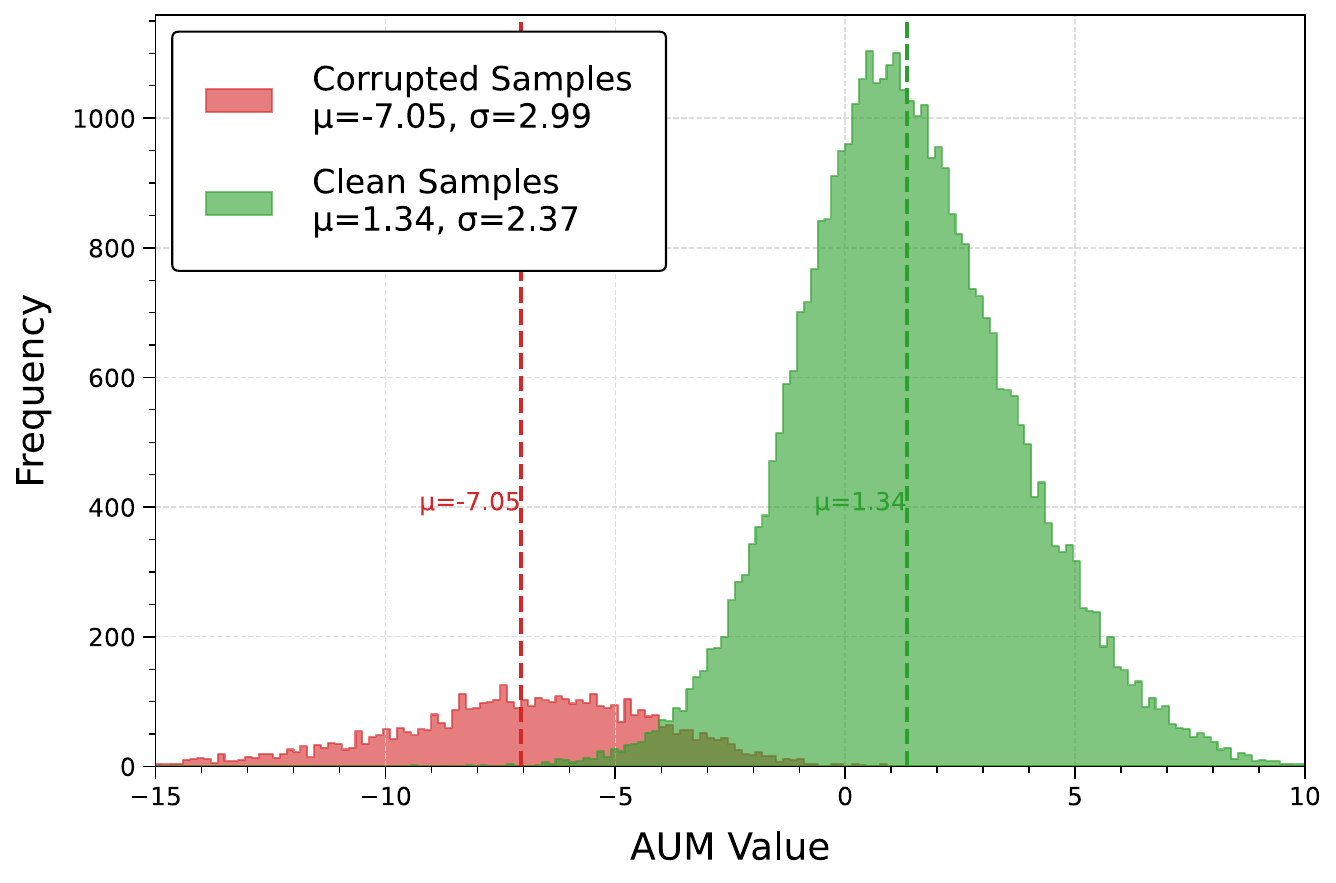}
        \caption{BL}
        \label{fig:aum_graph_BL}
    \end{subfigure}
    \hfill
    \begin{subfigure}[b]{0.32\textwidth}
        \centering
        \includegraphics[width=\linewidth]{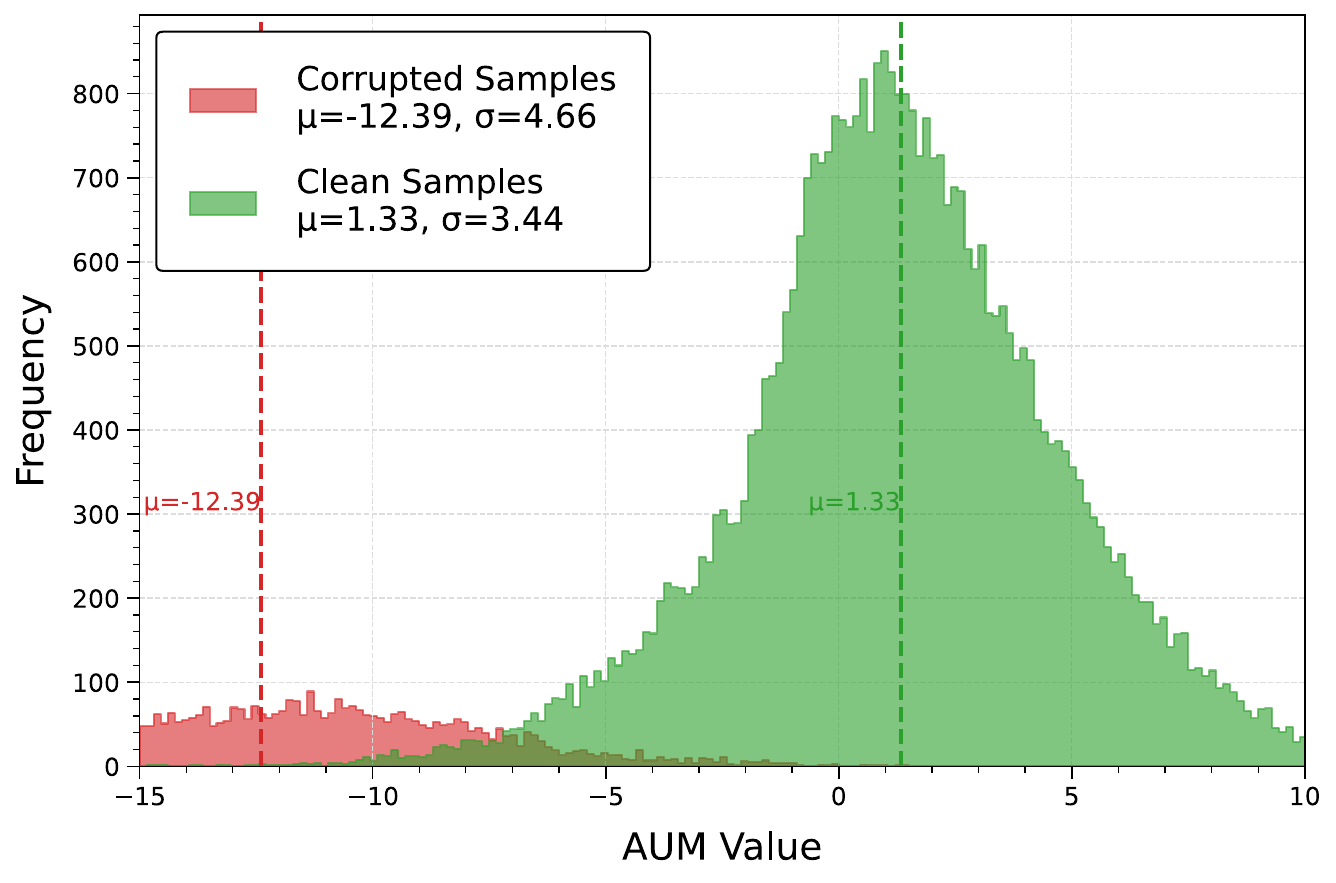}
        \caption{PZ}
        \label{fig:aum_graph_PZ}
    \end{subfigure}
    \caption{AUM distributions for artificially corrupted and clean samples in CIFAR-100 with corruption at $\eta=0.1$ for training with CE (\subref{fig:aum_graph_CE}), BL (\subref{fig:aum_graph_BL}), and PZ loss (\subref{fig:aum_graph_PZ}). Greater separation in distributions indicates a greater robustness of the loss function to mislabelled samples during training. Separation is measured using Cohen's d \{3.20 (CE) < 3.43 (BL) < 3.82 (PZ)\} and Wasserstein distance \{6.11 (CE) < 8.40 (BL) < 13.72 (PZ)\}.}
    \label{fig:training_dyn_AUM}
\end{figure}

As was explained in \Cref{sec:background}, the AUM statistic~\cite{pleiss2020identifying} measures the difference between the logit values for a training example's as-labelled class and its highest other class, averaged over the course of training.
Correctly labelled samples tend to exhibit positive AUM values, while mislabelled samples tend to have significantly lower AUM values due to conflicting gradient updates for samples of the same class. 
To better understand how the proposed loss functions affect the training dynamics of corrupt \vs clean data, the AUM is measured for all samples, and the AUM distributions for corrupt and clean data are compared. 
\Cref{fig:training_dyn_AUM} illustrates the AUM distributions for training with CE, BL, and PZ loss on the CIFAR-100 dataset with artificial corruption at $\eta=0.1$ (following the same method used throughout this paper).
The mean of each distribution is indicated using a dashed vertical line.
To quantitatively understand the impacts of training using the proposed loss functions, two measures of dissimilarity in distributions are computed: \mbox{Cohen’s d}~\cite{cohen2013statistical} (Effect size) and Wasserstein distance~\cite{kantorovich1960mathematical,vaserstein1969markov}, both of which are sensitive to differences in distribution position and width.
Note that an ideal robust training scheme would result in a large difference in the AUM distributions of corrupt and clean samples. 

Under CE training (\Cref{fig:aum_graph_CE}), clean and corrupted samples exhibit distinct but moderately overlapping distributions.
Although separation exists, the overlap reflects CE’s tendency to fit both clean and noisy labels, leading to partial memorization of mislabelled samples.

Training with Blurry Loss (\Cref{fig:aum_graph_BL}), produced a markedly improved separation. 
Quantitatively, a \mbox{Cohen’s d} of 3.43 (\vs 3.20 for CE), and Wasserstein distance of 8.40 (\vs 6.11 for CE) were measured, indicating improved separation of corrupt and clean data.
This increased separation of the distributions supports our intention and hypothesis that BL de-weights hard-to-classify (likely mislabelled) samples, reducing the extent to which the model fits to label errors.
Consequently, the discrimination between clean and noisy samples becomes more pronounced, enhancing the effectiveness of AUM-based label error detection.

The most dramatic effect was observed when using Piecewise-zero Loss (\Cref{fig:aum_graph_PZ}). 
Here, \mbox{Cohen’s d} increased to 3.82 (\vs 3.20 for CE), and the Wasserstein distance to 13.72 (\vs 6.11 for CE), indicating greater separation than both  CE and BL.
The sharp cutoff mechanism of PZ effectively prevents low-confidence samples from contributing to weight updates during training, almost eliminating their influence. 
As a result, mislabelled examples tend consistently to have highly negative AUM values, leading to the clearest distributional separation among all loss functions tested. 
Indeed, the F1 score notably increased by more than 13\% relative to CE, as was shown in \Cref{tab:aum_summary}.

Both BL and PZ loss functions substantially improve the margin-based separability of clean versus mislabelled samples compared to CE. 
These results validate our design objectives: the proposed loss functions mitigate the harmful fitting to noisy labels by selective de-weighting or exclusion.
This enhanced separability directly translates into improved performance for label error detection frameworks like AUM and Confident Learning.

\subsection{Expanded Versions of Results Shown in Main Paper}
\label{subsec:appendix_expanded}

\begin{figure}[h!]
    \centering
    \begin{subfigure}[b]{0.32\textwidth}
        \centering
        \includegraphics[width=1\linewidth]{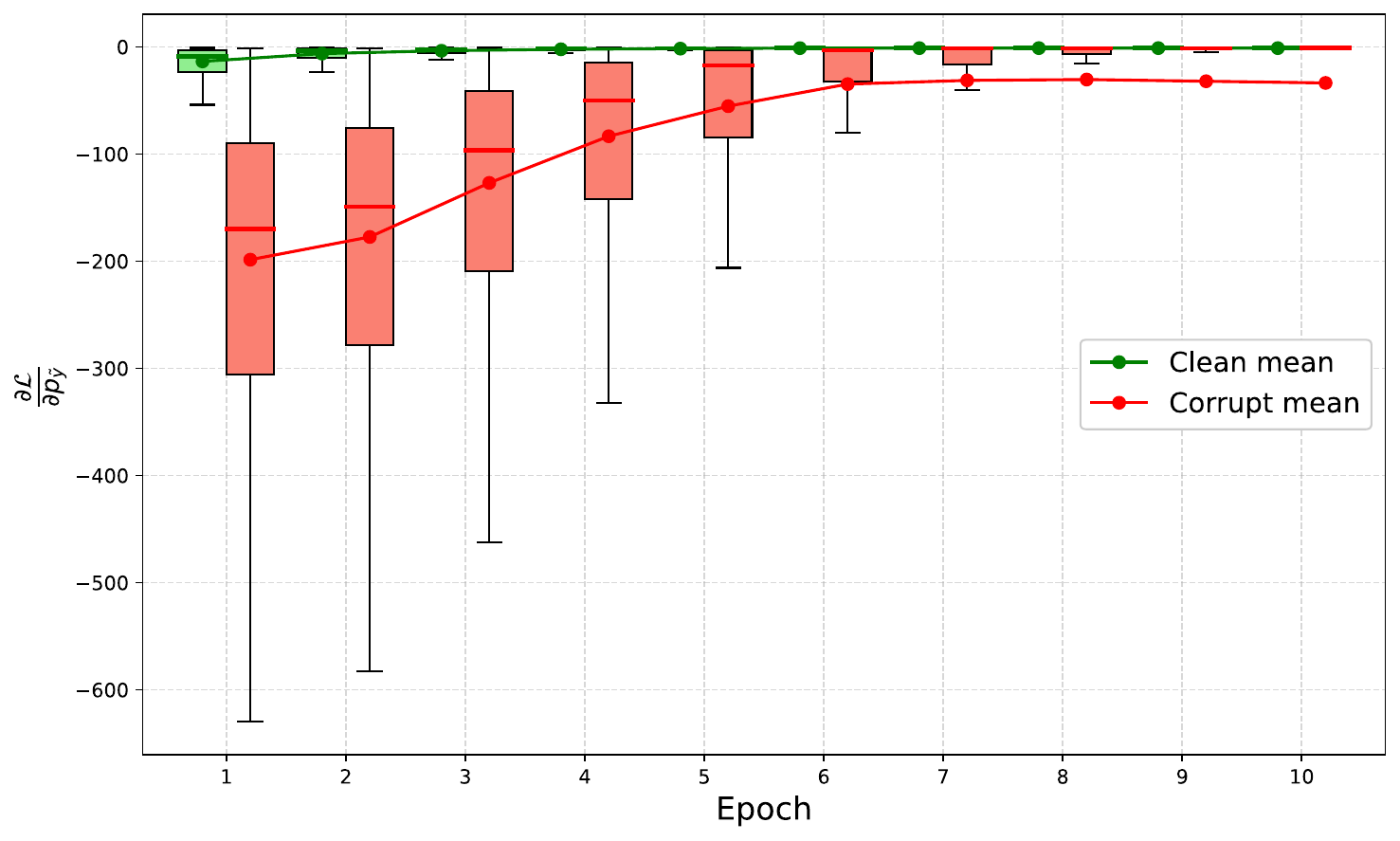}
        \caption{CE}
        \label{fig:gradients_Full_CE}
    \end{subfigure}
    \hfill
    \begin{subfigure}[b]{0.32\textwidth}
        \centering
        \includegraphics[width=1\linewidth]{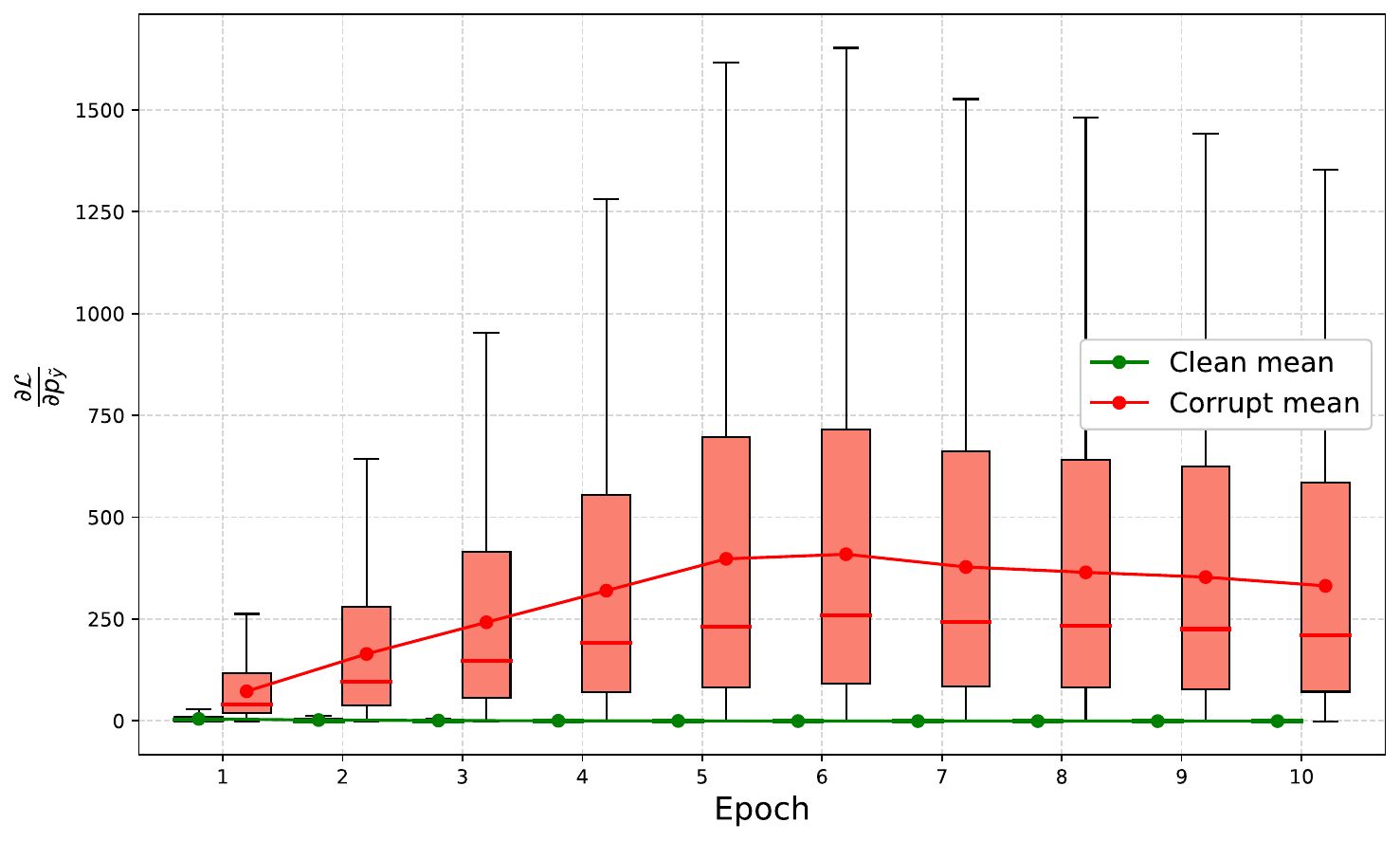}
        \caption{BL}
        \label{fig:gradients_Full_BL}
    \end{subfigure}
    \hfill
    \begin{subfigure}[b]{0.32\textwidth}
        \centering
        \includegraphics[width=\linewidth]{figures/gradients_PZ_CIFAR100_04.pdf}
        \caption{PZ}
        \label{fig:gradients_Full_PZ}
    \end{subfigure}
    \caption{A version of \Cref{fig:gradient_dist} without truncated vertical limits. Distributions of gradients of corrupt (red) and clean (green) data are plotted for each epoch for CIFAR-100 at $\eta=0.4$.}
    \label{fig:gradients_Full}
\end{figure}

\end{document}

%% file: figures/main_study_F1.tex
\begin{table}[tb]
  \centering
  \caption{Label error detections, averaged over 20 random trials, measured using the F1 score. The parameters ($\gamma$ and $c$) resulting in the best performance for each dataset and corruption rate are given for the proposed loss functions, Blurry Loss (BL) and Piecewise-zero Loss (PZ).}
  \label{tab:chart_df_F1}
  \resizebox{\textwidth}{!}{%
\begin{tabular}{c|c|ccccc|cccc}
\toprule
Dataset & $\eta$ & CE & FL & GCE & ANL-CE & ANL-FL & BL & $\gamma$ & PZ & \textit{c} \\
\midrule
\multirow{4}{*}{MNIST}   
 & 0.10 & 0.969±0.002 & 0.960±0.004 & 0.957±0.002 & 0.941±0.002 & 0.940±0.002 & \underline{\textbf{0.979±0.002}} & 0.3 & \textbf{0.974±0.001} & 0.005 \\
 & 0.20 & 0.975±0.001 & 0.967±0.002 & 0.975±0.001 & 0.970±0.001 & 0.969±0.001 & \underline{\textbf{0.984±0.001}} & 0.4 & \textbf{0.983±0.001} & 0.010 \\
 & 0.30 & 0.975±0.001 & 0.967±0.001 & 0.983±0.001 & 0.980±0.001 & 0.980±0.001 & \underline{\textbf{0.988±0.001}} & 0.4 & \textbf{0.986±0.001} & 0.020 \\
 & 0.40 & 0.973±0.001 & 0.963±0.002 & 0.987±0.001 & 0.985±0.000 & 0.985±0.001 & \underline{\textbf{0.990±0.000}} & 0.5 & \textbf{0.988±0.000} & 0.040 \\
\midrule

\multirow{4}{*}{Fashion-MNIST}
 & 0.10 & 0.824±0.003 & \underline{\textbf{0.831±0.004}} & 0.737±0.004 & 0.683±0.003 & 0.679±0.004 & \textbf{0.826±0.003} & 0.1 & 0.818±0.003 & 0.005 \\
 & 0.20 & 0.884±0.002 & 0.879±0.003 & 0.850±0.002 & 0.822±0.002 & 0.819±0.002 & \underline{\textbf{0.887±0.002}} & 0.2 & \textbf{0.885±0.003} & 0.005 \\
 & 0.30 & 0.907±0.002 & 0.899±0.001 & 0.900±0.001 & 0.881±0.002 & 0.879±0.001 & \underline{\textbf{0.917±0.001}} & 0.4 & \textbf{0.916±0.001} & 0.020 \\
 & 0.40 & 0.918±0.002 & 0.896±0.033 & 0.927±0.001 & 0.915±0.001 & 0.914±0.002 & \underline{\textbf{0.934±0.001}} & 0.5 & \textbf{0.933±0.001} & 0.040 \\
\midrule
\multirow{4}{*}{CIFAR-10}
 & 0.10 & 0.740±0.006 & 0.757±0.006 & 0.773±0.006 & 0.698±0.006 & 0.693±0.007 & \textbf{0.787±0.005} & 0.5 & \underline{\textbf{0.794±0.008}} & 0.040 \\
 & 0.20 & 0.773±0.004 & 0.789±0.004 & 0.858±0.004 & 0.821±0.004 & 0.817±0.005 & \textbf{0.860±0.004} & 0.6 & \underline{\textbf{0.863±0.004}} & 0.060 \\
 & 0.30 & 0.771±0.004 & 0.798±0.003 & 0.893±0.003 & 0.873±0.003 & 0.871±0.003 & \textbf{0.894±0.003} & 0.7 & \underline{\textbf{0.894±0.003}} & 0.080 \\
 & 0.40 & 0.769±0.003 & 0.801±0.003 & 0.909±0.003 & 0.901±0.002 & 0.899±0.003 & \textbf{0.912±0.003} & 0.8 & \underline{\textbf{0.912±0.003}} & 0.100 \\
\midrule
\multirow{4}{*}{CIFAR-100}
 & 0.10 & 0.503±0.009 & \underline{\textbf{0.520±0.009}} & 0.485±0.009 & 0.399±0.007 & 0.397±0.007 & 0.507±0.009 & 0.3 & \textbf{0.510±0.009} & 0.010 \\
 & 0.20 & 0.628±0.006 & 0.638±0.006 & 0.652±0.007 & 0.578±0.006 & 0.575±0.007 & \textbf{0.661±0.007} & 0.5 & \underline{\textbf{0.667±0.006}} & 0.020 \\
 & 0.30 & 0.680±0.006 & 0.688±0.006 & 0.742±0.006 & 0.682±0.006 & 0.680±0.006 & \textbf{0.744±0.006} & 0.6 & \underline{\textbf{0.748±0.006}} & 0.020 \\
 & 0.40 & 0.712±0.004 & 0.718±0.003 & 0.800±0.004 & 0.751±0.004 & 0.746±0.004 & \textbf{0.800±0.003} & 0.7 & \underline{\textbf{0.801±0.004}} & 0.020 \\

\bottomrule
\end{tabular}
  }
\end{table}

%% file: figures/main_study_BA.tex
\begin{table}[tb]
  \centering
  \caption{As in \Cref{tab:chart_df_F1}, but here reporting the Balanced Accuracy metric. The proposed BL and PZ perform strongly throughout.}
  \label{tab:chart_df_BA}
  \resizebox{\textwidth}{!}{%
\begin{tabular}{c|c|ccccc|cccc}
\toprule
Dataset & $\eta$ & CE & FL & GCE & ANL-CE & ANL-FL & BL & $\gamma$ & PZ & \textit{c} \\
\midrule
\multirow{4}{*}{MNIST}  
 & 0.10 & 0.980±0.002 & 0.971±0.004 & 0.992±0.001 & 0.992±0.000 & 0.992±0.000 & \underline{\textbf{0.993±0.001}} & 0.4 & \textbf{0.992±0.001} & 0.040 \\
 & 0.20 & 0.981±0.001 & 0.974±0.002 & 0.992±0.000 & 0.991±0.000 & 0.991±0.000 & \underline{\textbf{0.993±0.000}} & 0.5 & \textbf{0.993±0.000} & 0.020 \\
 & 0.30 & 0.980±0.001 & 0.971±0.001 & 0.992±0.000 & 0.991±0.000 & 0.990±0.000 & \underline{\textbf{0.993±0.000}} & 0.5 & \textbf{0.993±0.000} & 0.020 \\
 & 0.40 & 0.976±0.001 & 0.966±0.002 & 0.991±0.000 & 0.990±0.000 & 0.989±0.000 & \underline{\textbf{0.992±0.000}} & 0.5 & \textbf{0.991±0.000} & 0.040 \\
\midrule
\multirow{4}{*}{Fashion-MNIST}  
 & 0.10 & 0.942±0.002 & 0.941±0.002 & \textbf{0.949±0.002} & 0.943±0.001 & 0.942±0.001 & \underline{\textbf{0.951±0.001}} & 0.5 & 0.948±0.002 & 0.080 \\
 & 0.20 & 0.943±0.002 & 0.935±0.002 & 0.949±0.001 & 0.942±0.001 & 0.940±0.001 & \underline{\textbf{0.953±0.001}} & 0.4 & \textbf{0.950±0.001} & 0.040 \\
 & 0.30 & 0.939±0.001 & 0.929±0.001 & 0.948±0.001 & 0.939±0.001 & 0.938±0.001 & \underline{\textbf{0.953±0.001}} & 0.5 & \textbf{0.952±0.001} & 0.040 \\
 & 0.40 & 0.932±0.002 & 0.912±0.027 & 0.946±0.001 & 0.937±0.001 & 0.936±0.001 & \underline{\textbf{0.950±0.001}} & 0.5 & \textbf{0.949±0.001} & 0.040 \\
\midrule
\multirow{4}{*}{CIFAR-10}  
 & 0.10 & 0.934±0.002 & 0.931±0.002 & 0.955±0.002 & 0.946±0.002 & 0.945±0.002 & \underline{\textbf{0.956±0.001}} & 0.7 & \textbf{0.956±0.002} & 0.060 \\
 & 0.20 & 0.908±0.002 & 0.912±0.002 & 0.952±0.003 & 0.941±0.002 & 0.940±0.002 & \textbf{0.952±0.002} & 0.7 & \underline{\textbf{0.953±0.002}} & 0.080 \\
 & 0.30 & 0.862±0.003 & 0.879±0.002 & 0.944±0.002 & 0.934±0.002 & 0.933±0.002 & \textbf{0.945±0.002} & 0.7 & \underline{\textbf{0.945±0.002}} & 0.080 \\
 & 0.40 & 0.802±0.003 & 0.834±0.003 & 0.931±0.002 & 0.925±0.002 & 0.923±0.002 & \textbf{0.933±0.002} & 0.8 & \underline{\textbf{0.934±0.002}} & 0.100 \\
\midrule
\multirow{4}{*}{CIFAR-100}  
 & 0.10 & 0.833±0.004 & 0.829±0.004 & 0.853±0.004 & 0.827±0.005 & 0.825±0.005 & \textbf{0.855±0.004} & 0.5 & \underline{\textbf{0.856±0.004}} & 0.020 \\
 & 0.20 & 0.815±0.003 & 0.815±0.003 & 0.850±0.004 & 0.815±0.005 & 0.812±0.005 & \textbf{0.852±0.004} & 0.6 & \underline{\textbf{0.854±0.003}} & 0.020 \\
 & 0.30 & 0.785±0.004 & 0.789±0.005 & 0.844±0.005 & 0.800±0.005 & 0.797±0.006 & \textbf{0.844±0.004} & 0.7 & \underline{\textbf{0.846±0.005}} & 0.020 \\
 & 0.40 & 0.746±0.004 & 0.755±0.003 & 0.833±0.004 & 0.779±0.005 & 0.774±0.005 & \textbf{0.834±0.003} & 0.7 & \underline{\textbf{0.834±0.004}} & 0.020 \\

\bottomrule
\end{tabular}

  }
\end{table}

%% file: figures/pruned_CIFAR100.tex
\begin{table}[tb]
  \centering
  \caption{Label error detections on the CIFAR-100 cleaned test set, averaged over three random trials, measured using the F1 score ($\pm\sigma$). Performance is improved relative to \Cref{tab:chart_df_F1}, especially at lower $\eta$ and for PZ loss.}
  \label{tab:clean_CIFAR_results}
  % \resizebox{\textwidth}{!}{%
\scriptsize
\begin{tabular}{c|cccc}
\toprule
$\eta$ & CE & FL & BL & PZ \\
\midrule
0.1 & 0.576±0.027 & \textbf{0.587±0.030} & 0.580±0.024\,($\gamma$ = 0.3) & \underline{\textbf{0.590±0.020}}\,(\textit{c} = 0.020) \\
0.2 & 0.670±0.013 & 0.671±0.016 & \textbf{0.693±0.013}\,($\gamma$ = 0.3) & \underline{\textbf{0.717±0.017}}\,(\textit{c} = 0.015) \\
0.3 & 0.697±0.011 & 0.709±0.006 & \textbf{0.775±0.011}\,($\gamma$ = 0.5) & \underline{\textbf{0.780±0.013}}\,(\textit{c} = 0.020) \\
0.4 & 0.728±0.005 & 0.729±0.002 & \textbf{0.824±0.003}\,($\gamma$ = 0.7) & \underline{\textbf{0.830±0.007}}\,(\textit{c} = 0.025) \\
\bottomrule
\end{tabular}

  % }
\end{table}

%% file: figures/CIFAR-N_chart_horizontal_std.tex
\begin{table}[tb]
  \centering
  \caption{Label error detections on the CIFAR-10N and CIFAR-100N datasets with non-uniform corruption, averaged over five random trials, measured using the F1 score ($\pm\sigma$). The proposed BL and PZ loss functions continue to exceed baseline performance, with non-uniform corruption.}
  \label{tab:CIFAR_N_f1_comparison_std}
  \resizebox{\textwidth}{!}{%
\begin{tabular}{c|ccccc|cccc}
\toprule
Dataset & CE & Focal & GCE & ANL-CE & ANL-FL & BL & $\gamma$ & PZ & $c$ \\
\midrule
CIFAR-10N, $\eta \approx 0.1$ & 0.662 ± 0.004 & 0.634 ± 0.002 & 0.698 ± 0.002 & 0.652 ± 0.006 & 0.647 ± 0.003 & \textbf{0.698 ± 0.004} & 0.7 & \underline{\textbf{0.704 ± 0.006}} & 0.08 \\
CIFAR-100N, $\eta \approx 0.4$ & 0.700 ± 0.002 & 0.684 ± 0.002 & 0.741 ± 0.002 & 0.721 ± 0.002 & 0.721 ± 0.003 & \textbf{0.742 ± 0.003} & 0.8 & \underline{\textbf{0.745 ± 0.003}} & 0.10 \\
CIFAR-10N, $\eta \approx 0.4$ & 0.768 ± 0.001 & 0.769 ± 0.002 & 0.837 ± 0.002 & 0.865 ± 0.001 & 0.865 ± 0.002 & \underline{\textbf{0.867 ± 0.001}} & 0.9 & \textbf{0.866 ± 0.001} & 0.15 \\
\bottomrule
\end{tabular}

  }
\end{table}

%% file: figures/AUM_CIFAR-100.tex
\begin{table}[tb]
  \centering
  \caption{Label error detections using the AUM framework and with the CIFAR-100 dataset, averaged over three random trials, measured using the F1 score ($\pm\sigma$). The proposed losses continue to perform well using AUM, in addition to with CL (\Cref{tab:chart_df_F1,tab:chart_df_BA}).}
  \label{tab:aum_summary}
  \scriptsize
  \begin{tabular}{c|ccc}
    \toprule
   \multirow{2}{*}{\textbf{$\eta$}} & \multicolumn{3}{c}{\textbf{Loss Function}} \\ 
    \cmidrule(lr){2-4}
    %\textbf{Corruption Rate} 
      & CE 
      & BL ($\boldsymbol{\gamma=0.05}$) 
      & PZ (\textit{c} = 0.025) \\
    \midrule
    0.1 & 0.7452 ± 0.0071            & \textbf{0.7591 ± 0.0180}        & \underline{\textbf{0.8817 ± 0.0074}} \\
    0.2 & 0.8638 ± 0.0042            & \textbf{0.8692 ± 0.0032}        & \underline{\textbf{0.8903 ± 0.0085}} \\
    0.3 & \textbf{0.9019 ± 0.0021}   & \underline{\textbf{0.9076 ± 0.0031}} & 0.8914 ± 0.0044          \\
    0.4 & \textbf{0.9197 ± 0.0024}   & \underline{\textbf{0.9261 ± 0.0005}} & 0.8856 ± 0.0125          \\
    \bottomrule
  \end{tabular}
\end{table}